\documentclass[10pt,journal,compsoc]{IEEEtran}
\IEEEoverridecommandlockouts
\usepackage{microtype}
\usepackage{graphicx}
\usepackage{subfigure}
\usepackage{booktabs} % for professional tables
\usepackage{helvet} % Better Helvetica font support

\ifCLASSOPTIONcompsoc
  \usepackage[nocompress]{cite}
\else
  \usepackage{cite}
\fi

\usepackage{hyperref}
\hypersetup{
colorlinks=true,
linkcolor=blue,
anchorcolor=blue,
citecolor=blue}
\usepackage{enumerate}
\usepackage{caption}
\usepackage{subcaption}
\usepackage{pifont}
\usepackage{wrapfig}
\usepackage[utf8]{inputenc} % allow utf-8 input
\usepackage[T1]{fontenc}    % use 8-bit T1 fonts
\usepackage{url}            % simple URL typesetting
\usepackage{booktabs}       % professional-quality tables
\usepackage{amsfonts}       % blackboard math symbols
\usepackage{nicefrac}       % compact symbols for 1/2, etc.
\usepackage{microtype}      % microtypography
\usepackage{xcolor}         % colors
\usepackage{tikz}

\usepackage{subfigure}
\usepackage{enumitem}
\usepackage{rotating}
\usepackage{adjustbox}
\usepackage{amsmath,amssymb,amsfonts}
\usepackage{amsthm}

\usepackage{cleveref}
\usepackage{multirow}
\usepackage{booktabs,bm}
\usepackage{algorithm}
\usepackage{algorithmic}
\usepackage{colortbl}
\usepackage{tcolorbox}

\definecolor{grey}{rgb}{0.89,0.71,0.57}
\definecolor{pink}{rgb}{1,0.94,1}
\definecolor{purple}{rgb}{0.84,0.78,1}
\definecolor{white}{rgb}{1,1,1}

\newcommand{\insightbox}[1]{%
    \begin{tcolorbox}[colframe=black!70, colback=yellow!5, boxrule=1pt, arc=2mm]
        \includegraphics[width=0.3cm]{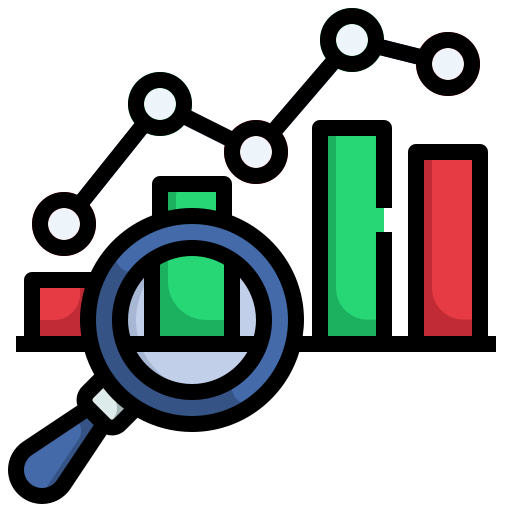}
        \textbf{\small#1}
    \end{tcolorbox}
}

\def\method{\textsc{UniFer}}
\def\benchmark{\textsc{FerBench}}
\begin{document}

\title{Rethinking Facial Expression Recognition in the Era of Multimodal Large Language Models: Benchmark, Datasets, and Beyond}

\author{Fan Zhang, Haoxuan Li, Shengju Qian$^{\dagger}$, Xin Wang, Zheng Lian, Hao Wu, Zhihong Zhu, Yuan Gao,\\ Qiankun Li, Yefeng Zheng,~\IEEEmembership{Fellow,~IEEE}, Zhouchen Lin$^{\dagger}$,~\IEEEmembership{Fellow,~IEEE}, Pheng-Ann Heng

\IEEEcompsocitemizethanks{
\IEEEcompsocthanksitem $^{\dagger}$Corresponding authors: Shengju Qian, Zhouchen Lin.
\IEEEcompsocthanksitem Fan Zhang and Pheng-Ann Heng are with The Chinese University of Hong Kong, Hong Kong, China. (e-mail: fzhang@link.cuhk.edu.hk)
\IEEEcompsocthanksitem Haoxuan Li and Zhouchen Lin are with Peking University, Beijing, China.
\IEEEcompsocthanksitem Shengju Qian, Xin Wang, and Zhihong Zhu are with Tencent, Shenzhen, China.
\IEEEcompsocthanksitem Zheng Lian is with Institute of Automation, Chinese Academy of Sciences, Beijing, China.
\IEEEcompsocthanksitem Hao Wu and Yuan Gao are with Tsinghua University, Beijing, China.
\IEEEcompsocthanksitem Qiankun Li is with Nanyang Technological University, Singapore, SG.
\IEEEcompsocthanksitem Yefeng Zheng is with Westlake University, Hangzhou, China.
}
%\thanks{Manuscript received 07 September 2022.}
}

\IEEEtitleabstractindextext{%
\begin{abstract}

Multimodal Large Language Models (MLLMs) have revolutionized numerous research fields, including computer vision and affective computing. As a pivotal challenge in this interdisciplinary domain, facial expression recognition (FER) has evolved from separate, domain-specific models to more unified approaches. One promising avenue to unify FER tasks is converting conventional FER datasets into visual question-answering (VQA) formats, enabling the direct application of powerful generalist MLLMs for inference. However, despite the success of cutting-edge MLLMs in various tasks, their performance on FER tasks remains largely unexplored. To address this gap, we provide \benchmark{}, a systematic benchmark that incorporates 20 state-of-the-art MLLMs across four widely used FER datasets. Our results reveal that, while MLLMs exhibit good classification performance, they still face significant limitations in reasoning and interpretability. To this end, we introduce post-training strategies aimed at enhancing the facial expression reasoning capabilities of MLLMs. Specifically, we curate two high-quality and large-scale datasets: \method{}-CoT-230K for cold-start initialization and \method{}-RLVR-360K for reinforcement learning with verifiable rewards (RLVR), respectively. Building upon them, we develop a unified and interpretable FER foundation model termed \method{}-7B, which outperforms many open-sourced and closed-source generalist MLLMs (\emph{e.g.}, Gemini-2.5-Pro and Qwen2.5-VL-72B).
Our source code and curated datasets are available at \url{https://github.com/zfkarl/UniFER}.
\end{abstract}

}
\maketitle

\section{Introduction}

\IEEEPARstart{F}{acial} expression recognition (FER) \cite{li2020deep,tian2011facial,kumari2015facial} constitutes a long-standing and fundamental problem in the domains of affective computing and computer vision. The primary objective is to automatically discern human emotions from facial features, often leveraging visual clues such as action units \cite{mao2025facial} and muscle movements \cite{zhang2011facial}. This task bears significant importance across a diverse spectrum of applications, including human-computer interaction \cite{chattopadhyay2020facial,cowie2001emotion}, emotionally responsive digital avatars \cite{kakarla2014real,yang2011facial}, and diagnostic support in healthcare and psychological well-being \cite{tacconi2008activity,pepa2021automatic}.

\begin{figure}[t]
    \centering
    \includegraphics[width=\linewidth]{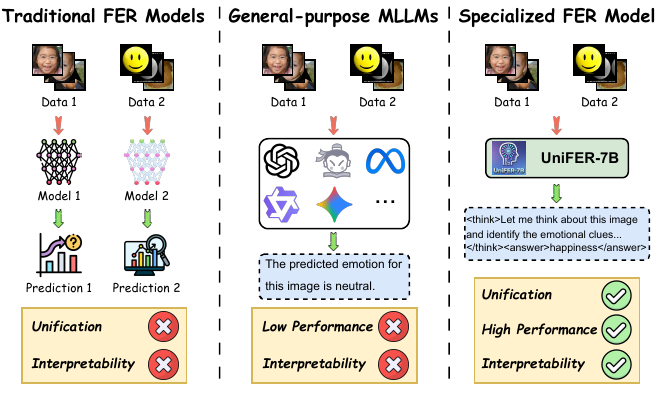}
    \caption{An illustration of traditional FER models, general-purpose MLLMs, and our proposed specialized FER model.}
    \label{fig:intro}
    \vspace{-4mm}
\end{figure}

Prior to the era of multimodal large language models (MLLMs), numerous efforts have been devoted to extracting more discriminative visual features to improve emotion classification performance.
For instance, convolutional neural network (CNN)-based models \cite{simonyan2014very,he2016deep} utilize convolutional and pooling layers to effectively capture both global and local features from facial images, making them well-suited for emotion recognition tasks.
Meanwhile, transformer-based models \cite{dosovitskiy2020image,liu2021swin} employ attention mechanisms that are particularly adept at modeling dynamic relationships across spatial regions and channels, thereby enhancing the ability to distinguish fine-grained emotion categories.
While these models demonstrate impressive performance, their limitations remain noteworthy (Fig. \ref{fig:intro} Left).
\ding{182} First, the common practice of projecting emotion labels into one-hot vectors for training such black-box models leads to a loss of semantic information \cite{zhang2021relative,zeng2022face2exp}. As a result, the models learn merely discriminative representations without the ability to interpret the reasoning behind their predictions.
\ding{183} Second, owing to the inherent domain discrepancies among different FER datasets \cite{zhang2022learn,chen2021understanding}, it is often necessary to train separate models tailored to each specific dataset, rather than establishing a unified foundation model adaptable to all FER tasks. This approach suffers from poor scalability, increasing both the difficulty and cost of model deployment.
Therefore, a fundamental and pressing question arises:

\insightbox{How can we establish a unified and interpretable paradigm for Facial Expression Recognition (FER)?}

With the recent advancement of MLLMs, this long-standing challenge shows significant potential for effective resolution, as MLLMs inherently possess strengths in task adaptability, scalability, and interpretability \cite{lian2024gpt,feng2025video}.
By reformulating traditional FER datasets into visual question answering (VQA) formats—a paradigm naturally suited to general-purpose MLLMs—these models can be effectively applied to emotion understanding tasks. Consequently, it becomes feasible to leverage off-the-shelf MLLMs for unified inference across multiple FER datasets under a consistent framework, thereby overcoming previous limitations.

To this end, we propose \benchmark{}, the first-ever comprehensive benchmark for evaluating the emotional intelligence of MLLMs in FER tasks.
In particular, we collect 11K facial images along with their annotated labels from four widely-used FER datasets \cite{li2017reliable,li2019reliable,BarsoumICMI2016,mollahosseini2017affectnet,zhang2024generalizable}. Each sample is reformatted into a VQA format by embedding the emotion label into a consistent, predefined prompt template. We subsequently carry out a systematic evaluation of 20 cutting-edge MLLMs across our benchmark. To ensure a fair comparison, all models are tested under the same prompt formulations and temperature settings during the inference process.
Taking a closer look at the emotional intelligence of MLLMs, we observe that while they can achieve competitive prediction accuracy, they often treat FER merely as a classification problem and lack the ability to provide reasonable, explanatory rationales for their predictions (Fig. \ref{fig:intro} Mid).

Inspired by the recent success of large reasoning models (LRMs) \cite{xu2025towards,guo2025deepseek,patil2025advancing}, we propose leveraging post-training techniques to further enhance the understanding and reasoning capabilities of MLLMs for FER tasks. As FER provides verifiable ground-truth answers, we can employ reinforcement learning with verifiable rewards (RLVR) for effective model training.
Prior to RL training, a cold-start initialization process is crucial for equipping the model with a preliminary rollout capability \cite{feng2025video}. Corresponding to these two phases, we collect two large-scale datasets from public sources: \method{}-RLVR-360K and \method{}-CoT-230K. The former comprises 360K facial images and corresponding text-based QA pairs. To ensure generalization, we maintain a uniform answer template while diversifying the question formulations using LLMs, thereby preventing overfitting to fixed patterns. The latter is a high-quality Chain-of-Thought (CoT) dataset and also constitutes a subset of the former. We synthesize long CoT reasoning trajectories through rule-based injection and LLM-based generation, followed by a multi-stage quality control process to filter out low-quality samples, ensuring high efficacy during cold-start training.

Leveraging these two curated high-quality datasets, we employ a two-stage training framework for the baseline model Qwen2.5-VL-7B \cite{Qwen2.5-VL}, which involves standard supervised fine-tuning (SFT) followed by group relative policy optimization (GRPO). This post-training process yields a specialized FER foundation model, named \method{}-7B.
Further experimental results highlight the advantages of \method{}-7B in three key aspects (Fig. \ref{fig:intro} Right):
\ding{182} \textbf{Unification}—\method{}-7B enables consistent modeling, training, and inference across multiple FER datasets, serving as a one-for-all FER foundation model.
\ding{183} \textbf{High Performance}—Under both task-level and category-level evaluations, \method{}-7B establishes new state-of-the-art (SOTA) performance. It not only surpasses larger open-source models (\emph{e.g.}, Qwen2.5-VL-72B \cite{Qwen2.5-VL} and InternVL3-78B \cite{zhu2025internvl3}) but also outperforms leading closed-source models (\emph{e.g.}, GPT-5 \cite{openaigpt5} and Gemini-2.5-Pro \cite{gemini25pro}).
\ding{184} \textbf{Interpretability}—\method{}-7B provides complete reasoning trajectories that reveal the rationale behind its predictions, while also demonstrating advanced higher-order reasoning abilities, such as verification and self-reflection. This marks the first emergence of the “aha moment” in the FER domain.

In summary, the main contributions of this paper are threefold:
\begin{itemize}[leftmargin=*]
    \item \textbf{\textit{Systematic Benchmark}.} We introduce \benchmark{}, the first-ever comprehensive benchmark specifically designed to evaluate the emotional intelligence of MLLMs in FER tasks. Through systematic assessments of 20 cutting-edge MLLMs on 11K facial images, we reveal both their strengths and limitations, paving the way for future research on MLLM-based FER and affective computing.
    \item \textbf{\textit{Meticulous Datasets}.} We curate two large-scale and high-quality datasets, \method{}-CoT-230K and \method{}-RLVR-360K, designed for the SFT and RLVR stages of post-training in FER tasks, respectively. These datasets serve as a robust foundation for future research and can be seamlessly integrated into the training process of any MLLMs.
    \item \textbf{\textit{FER Foundation Model}.} Going beyond this, we introduce \method{}-7B, an all-in-one FER foundation model that features unification, high performance, and interpretability. Experimental results across multiple datasets demonstrate that \method{}-7B outperforms both SOTA closed-source and open-source MLLMs, setting a new standard in this field.
\end{itemize}

\section{Related Work}

\begin{figure*}[t]
    \centering
    \includegraphics[width=\linewidth]{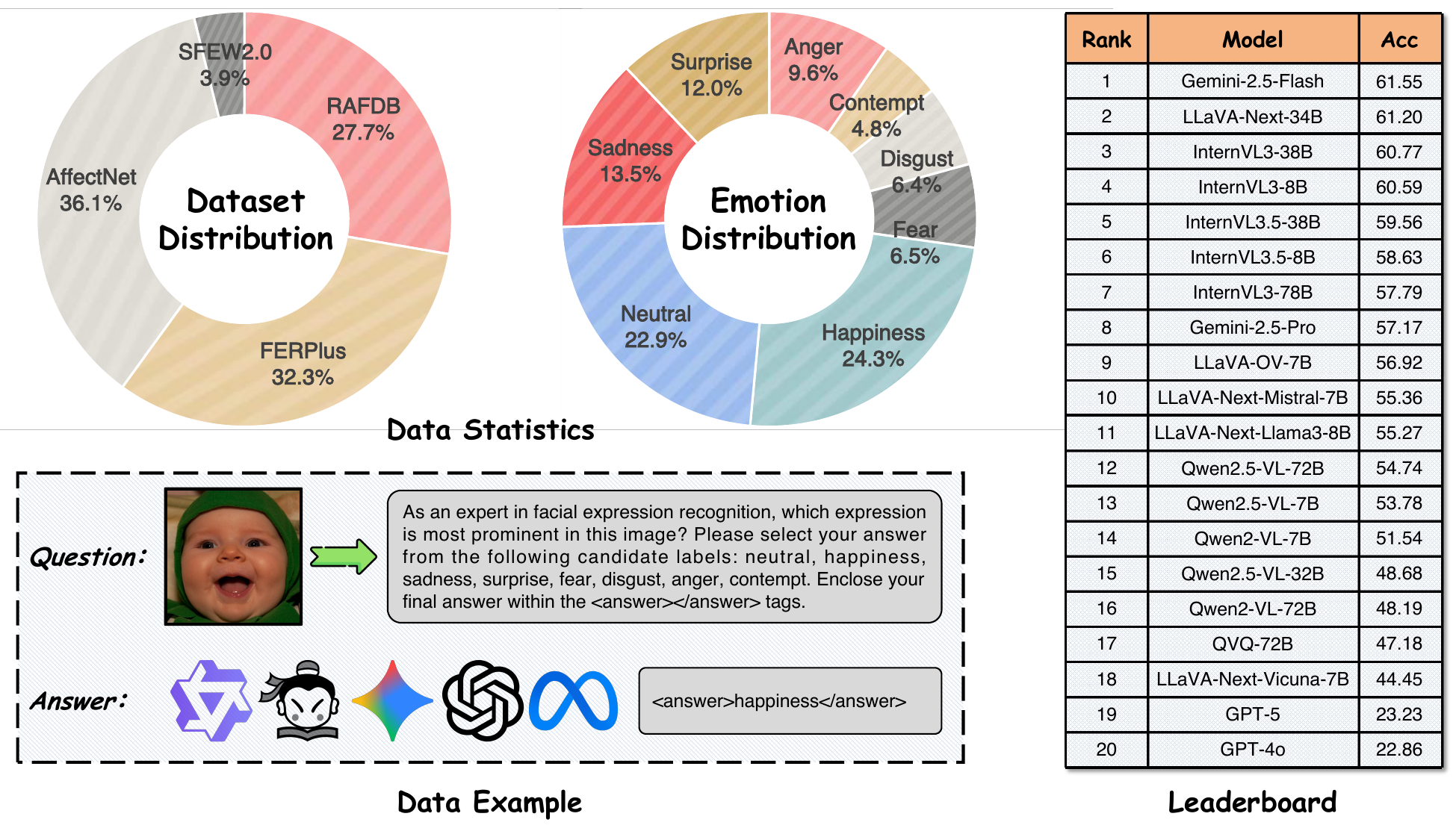}
    \caption{An overview of our proposed \benchmark{}. We incorporate 11K facial images and 20 cutting-edge MLLMs for open and fair evaluation. The top-performing model (\emph{i.e.}, Gemini-2.5-Flash) only achieves $61.55\%$ accuracy on \benchmark{}.}
    \label{fig:bench}
    %\vspace{-4mm}
\end{figure*}

\subsection{Facial Expression Recognition}

Facial Expression Recognition (FER) \cite{revina2021survey,bettadapura2012face,huang2019facial} is a core task at the intersection of computer vision and affective computing, with the primary goal of accurately identifying human emotions from facial clues. Prior to the emergence of multimodal large language models (MLLMs), research in FER mainly focus on extracting high-quality visual features, ranging from handcrafted descriptors \cite{ng2003sift,dalal2005histograms,shan2009facial,liu2002gabor,hu2008multi,luo2013facial,pietikainen2011computer} to learning-based representations \cite{li2017reliable,cheng2023semi,she2021dive,wang2020suppressing,zhang2024leaf,xue2021transfer}. Once features are obtained, they can be typically fed into supervised classifiers, such as support vector machines (SVMs), softmax layers, or logistic regression, to predict categorical emotion labels.
With the advent of MLLMs, and their remarkable performance on visual question answering (VQA) tasks \cite{dai2023instructblip,liu2023visual,huang2025visual}, a new line of research has emerged that explores leveraging these models for FER. This paradigm shift has attracted growing attention, as MLLMs open up the possibility of addressing FER in a more unified and generalizable manner.

\subsection{Multimodal Large Language Models}

Benefiting from the success of large language models (LLMs) \cite{ouyang2022training,vicuna2023,liu2024deepseek,guo2025deepseek}, significant progress has also been made in multimodal understanding tasks. The latest paradigm involves using multimodal large language models (MLLMs) composed of three key components: a vision encoder, a connector, and an LLM, for task processing. The vision encoder is responsible for extracting visual features, with mainstream architectures including large-scale image-text pre-trained models such as CLIP \cite{radford2021learning} and SigLIP \cite{zhai2023sigmoid}. The connector serves as a bridge between visual and linguistic representation spaces, enabling vision-language alignment. Commonly used connector structures range from simple yet effective multilayer perceptrons (MLPs) \cite{liu2023visual} to more complex designs like Q-Former \cite{dai2023instructblip,li2023blip}. The LLM component is tasked with comprehending visual semantics and customized textual instructions, subsequently generating structured textual outputs as task responses. Powerful LLM backbones, including closed-source models like the GPT \cite{brown2020language} and Gemini \cite{team2023gemini} series, as well as open-source alternatives such as Qwen \cite{bai2023qwen} and LLaMA \cite{touvron2023llama}, can be readily integrated into MLLMs.

\subsection{Affective Computing with MLLMs}

Affective computing is a long-standing task in artificial intelligence and holds significant potential for integration with multimodal large language models (MLLMs). Mainstream MLLMs are typically evaluated across a variety of visual question-answering (VQA) tasks, including conversational reasoning \cite{li2024llavanext-strong,lu2024wildvision}, general knowledge comprehension \cite{fu2024mmecomprehensiveevaluationbenchmark,liu2024mmbench}, optical character recognition (OCR) \cite{chen2021websrc,liu2023hidden}, mathematical problem-solving \cite{lu2023mathvista,zhang2024mathverse}, hallucination mitigation \cite{sun2023aligning,li2023evaluating}, and video understanding \cite{maaz2023video,fu2025video}.
In contrast, affective computing remains a relatively underexplored yet highly impactful domain, with broad applications in human-computer interaction \cite{pantic2005affective,hudlicka2003feel}, robotics \cite{gervasi2023applications,devillers2021human}, healthcare \cite{jin2024affective,yannakakis2018enhancing}, and education \cite{wu2016review,yadegaridehkordi2019affective}. Although some recent efforts have begun to integrate MLLMs with multimodal emotion recognition \cite{lian2024affectgpt,zhang2025mme,lian2025emoprefer,yang2025humanomniv2,zhao2025r1}, these approaches often focus on holistic contextual clues from individuals rather than emphasizing fine-grained facial expression details—as is central to facial expression recognition (FER).

\section{The Proposed Benchmark: \benchmark{}}

In Sec. \ref{sec:dataset}, we first present our data collection and transformation strategy, followed by a description of the experimental settings in Sec. \ref{sec:setting}. We then conduct an in-depth performance analysis of the the evaluated MLLMs in Sec. \ref{sec:analysis}.

\begin{table*}[t]
\centering
\caption{Task-level comparisons (in $\%$) across various MLLMs on \benchmark{}. Best results are marked in \textbf{bold}.}
\label{tab:main_cmp}
\resizebox{\linewidth}{!}{%
% \begin{tabular}{lcccccccccc}
\begin{tabular}{l@{\hspace{5mm}}c@{\hspace{5mm}}c@{\hspace{5mm}}c@{\hspace{5mm}}c@{\hspace{5mm}}c@{\hspace{5mm}}c@{\hspace{5mm}}c@{\hspace{5mm}}c@{\hspace{5mm}}c@{\hspace{5mm}}c}
\toprule[1.2pt]
\multirow{2.5}{*}{\textbf{Model}} & \multicolumn{2}{c}{~\hspace{-7mm}\textbf{RAFDB}} & \multicolumn{2}{c}{~\hspace{-7mm}\textbf{FERPlus}} & \multicolumn{2}{c}{~\hspace{-7mm}\textbf{AffectNet}} & \multicolumn{2}{c}{~\hspace{-7mm}\textbf{SFEW2.0}} & \multicolumn{2}{c}{~\hspace{-5mm}\textbf{Overall}} \\
 \cmidrule{2-11}& \textbf{Acc} & \textbf{F1} & \textbf{Acc} & \textbf{F1} & \textbf{Acc} & \textbf{F1} & \textbf{Acc} & \textbf{F1} & \textbf{Acc} & \textbf{F1} \\
\midrule
LLaVA-Next-Vicuna-7B \cite{liu2023improved} & 55.96 & 39.36 & 48.61 & 33.41 & 32.78 & 27.41 & 36.19 & 28.75 & 44.45 & 30.27 \\
LLaVA-Next-Mistral-7B \cite{liu2023improved} & 72.49 & 59.02 & 66.25 & 42.55 & 34.05 & 26.10 & 41.07 & 28.15 & 55.36 & 34.84 \\
LLaVA-Next-Llama3-8B \cite{li2024llavanext-strong} & 63.49 & 45.94 & \textbf{72.29} & 39.61 & 35.55 & 28.67 & 38.52 & 28.04 & 55.27 & 36.15 \\
LLaVA-OV-7B \cite{li2024llava} & 64.50 & 49.49 & 68.63 & 47.46 & 41.63 & 30.05 & 47.80 & \textbf{39.01} & 56.92 & 36.29 \\
Qwen2-VL-7B \cite{Qwen2-VL} & 56.58 & 45.68 & 68.99 & 43.09 & 32.93 & 28.73 & 43.62 & 33.38 & 51.54 & 37.67 \\
Qwen2.5-VL-7B \cite{Qwen2.5-VL} & 62.68 & 50.26 & 67.53 & 46.25 & 35.68 & 26.97 & 44.55 & 36.22 & 53.78 & 35.34 \\
InternVL3-8B \cite{zhu2025internvl3} & 74.05 & 52.69 & 66.72 & 43.34 & 46.15 & 39.33 & 47.80 & 37.27 & 60.59 & 44.54 \\
InternVL3.5-8B \cite{wang2025internvl3_5} & 73.99 & 50.92 & 67.90 & 41.60 & 39.95 & 31.02 & 45.94 & 31.81 & 58.63 & 39.28 \\
\midrule
LLaVA-Next-34B \cite{liu2023improved} & 77.93 & \textbf{60.56} & 71.26 & \textbf{48.43} & 40.80 & 34.92 & \textbf{48.03} & 37.76 & 61.20 & 44.36 \\
Qwen2.5-VL-32B \cite{Qwen2.5-VL} & 54.11 & 48.32 & 61.04 & 41.58 & 34.13 & 29.89 & 42.69 & 34.79 & 48.68 & 36.97 \\
InternVL3-38B \cite{zhu2025internvl3} & \textbf{78.68} & 56.98 & 66.25 & 43.76 & 43.90 & 35.10 & 44.55 & 35.05 & 60.77 & 42.61 \\
InternVL3.5-38B \cite{wang2025internvl3_5} & 76.76 & 54.68 & 68.68 & 41.62 & 40.18 & 31.85 & 41.53 & 34.58 & 59.56 & 40.15 \\
\midrule
Qwen2-VL-72B \cite{Qwen2-VL} & 50.07 & 44.22 & 66.97 & 41.05 & 30.80 & 27.22 & 40.60 & 31.23 & 48.19 & \textbf{53.64} \\
Qwen2.5-VL-72B \cite{Qwen2.5-VL} & 66.10 & 53.92 & 69.16 & 46.12 & 34.28 & 30.74 & 44.32 & 35.91 & 54.74 & 41.00 \\
QVQ-72B \cite{qvq-72b-preview} & 50.42 & 37.03 & 61.04 & 36.28 & 33.33 & 27.68 & 37.82 & 30.46 & 47.18 & 32.34 \\
InternVL3-78B \cite{zhu2025internvl3} & 72.69 & 54.15 & 62.83 & 40.27 & 43.50 & 35.83 & 42.69 & 33.67 & 57.79 & 41.43 \\
\midrule
GPT-4o \cite{openaigpt4o} & 22.88 & 7.11 & 34.59 & 6.57 & 12.73 & 2.96 & 19.49 & 4.11 & 22.86 & 5.00 \\
GPT-5 \cite{openaigpt5} & 23.08 & 9.35 & 35.07 & 11.66 & 13.30 & 4.68 & 18.33 & 6.71 & 23.23 & 7.34 \\
Gemini-2.5-Flash \cite{gemini25flash} & 72.98 & 55.60 & 68.32 & 44.95 & 48.30 & \textbf{45.38} & 47.10 & 37.20 & \textbf{61.55} & 45.47 \\
Gemini-2.5-Pro \cite{gemini25pro} & 66.75 & 50.95 & 57.99 & 39.78 & \textbf{50.53} & 43.11 & 43.85 & 36.33 & 57.17 & 44.29 \\
\bottomrule[1.2pt]
\end{tabular}%
}
%\vspace{-2mm}
\end{table*}

\begin{table*}[ht]
\centering
\caption{Category-level comparisons (in $\%$) across various MLLMs on \benchmark{}. Best results are marked in \textbf{bold}.}
\label{tab:cate_cmp}
\resizebox{\linewidth}{!}{%
\begin{tabular}{lccccccccc}
\toprule[1.2pt]
\textbf{Model} & \textbf{Anger} & \textbf{Contempt} & \textbf{Disgust} & \textbf{Fear} & \textbf{Happiness} & \textbf{Neutral} & \textbf{Sadness} & \textbf{Surprise} & \textbf{Avg} \\
\midrule
LLaVA-Next-Vicuna-7B \cite{liu2023improved} & 31.74 & 0.00 & 27.10 & 6.66 & 83.85 & 6.26 & 47.67 & 38.91 & 30.27 \\
LLaVA-Next-Mistral-7B \cite{liu2023improved} & 39.94 & 0.00 & 20.48 & 8.81 & 84.82 & 63.41 & 47.91 & 48.16 & 39.19 \\
LLaVA-Next-Llama3-8B \cite{li2024llavanext-strong} & 48.29 & 0.00 & 38.85 & 0.28 & 82.06 & 61.90 & 6.31 & 51.54 & 36.15 \\
LLaVA-OV-7B \cite{li2024llava} & 51.10 & 0.37 & 14.24 & 9.65 & 78.98 & 54.98 & 58.36 & 58.91 & 40.83 \\
Qwen2-VL-7B \cite{Qwen2-VL}  & 46.97 & 0.37 & 22.26 & 3.28 & 67.29 & 54.60 & 53.91 & 52.70 & 37.67 \\
Qwen2.5-VL-7B \cite{Qwen2.5-VL} & 47.99 & 5.59 & 14.63 & 4.87 & 74.89 & 59.68 & 55.84 & 54.58 & 39.76 \\
InternVL3-8B \cite{zhu2025internvl3} & \textbf{54.41} & 14.55 & 38.31 & 41.35 & 85.56 & \textbf{65.08} & 61.57 & 40.02 & 50.11 \\
InternVL3.5-8B \cite{wang2025internvl3_5} & 45.93 & 5.54 & 38.34 & 14.75 & 82.05 & 63.64 & 44.87 & 58.37 & 44.19 \\
\midrule
LLaVA-Next-34B \cite{liu2023improved} & 51.22 & 0.36 & 35.79 & 3.01 & \textbf{86.11} & 60.62 & \textbf{64.00} & 53.79 & 44.36 \\
Qwen2.5-VL-32B \cite{Qwen2.5-VL} & 37.26 & 4.90 & 12.83 & 22.92 & 60.77 & 53.97 & 51.17 & 51.94 & 36.97 \\
InternVL3-38B \cite{zhu2025internvl3} & 54.03 & 8.53 & 32.02 & 23.24 & 83.97 & 58.36 & 63.37 & 59.95 & 47.93 \\
InternVL3.5-38B \cite{wang2025internvl3_5} & 49.40 & 3.36 & 33.36 & 17.42 & 82.68 & 61.50 & 56.56 & 57.08 & 45.17 \\
\midrule
Qwen2-VL-72B \cite{Qwen2-VL} & 45.21 & 0.00 & 14.27 & 6.43 & 56.54 & 51.61 & 52.26 & 54.08 & 35.05 \\
Qwen2.5-VL-72B \cite{Qwen2.5-VL} & 48.61 & 0.68 & 16.32 & 22.15 & 73.55 & 56.77 & 52.51 & 57.40 & 41.00 \\
QVQ-72B \cite{qvq-72b-preview} & 39.19 & 1.74 & 7.22 & 7.85 & 70.96 & 50.09 & 36.76 & 44.88 & 32.34 \\
InternVL3-78B \cite{zhu2025internvl3} & 51.25 & 8.15 & 29.95 & 30.13 & 83.34 & 54.35 & 60.04 & 55.63 & 46.61 \\
\midrule
GPT-4o \cite{openaigpt4o} & 2.22 & 0.00 & 0.28 & 0.82 & 1.40 & 37.06 & 1.45 & 1.77 & 5.63 \\
GPT-5 \cite{openaigpt5} & 2.93 & 0.00 & 0.82 & 4.41 & 16.33 & 35.90 & 3.09 & 2.54 & 8.25 \\
Gemini-2.5-Flash \cite{gemini25flash} & 48.86 & 9.14 & \textbf{41.82} & \textbf{42.52} & 81.53 & 62.25 & 59.69 & \textbf{63.43} & \textbf{51.15} \\
Gemini-2.5-Pro \cite{gemini25pro} & 46.44 & \textbf{20.49} & 39.66 & 42.42 & 81.22 & 47.43 & 57.78 & 63.16 & 49.83\\
\bottomrule[1.2pt]
\end{tabular}%
}
\end{table*}

\subsection{Data Collection and Transformation}
\label{sec:dataset}

We utilize four classic and widely adopted FER datasets, RAFDB \cite{li2017reliable,li2019reliable}, FERPlus \cite{BarsoumICMI2016}, AffectNet \cite{mollahosseini2017affectnet}, and SFEW2.0 \cite{zhang2024generalizable}, as our source data. To prevent potential data leakage, we collect images exclusively from the test sets, resulting in a total of 11,072 images for VQA transformation. The distributions of datasets and emotion categories are provided in Fig. \ref{fig:bench}. For each facial image and its corresponding emotion label, we perform format conversion using a predefined prompt template. Since current MLLMs still exhibit limited recognition capability in open-set scenarios, we opt for a closed-set setting that provides all candidate labels in the prompt template. In the system prompt, we instruct the evaluated MLLM to act as an expert in FER to enhance its understanding of the task. In addition, we employ a consistent prompt strategy to ensure that all evaluated MLLMs remain unaffected by prompt design variations. 
%A sample of the converted data is illustrated in Fig. \ref{fig:bench}.

%\vspace{-1mm}
\subsection{Experimental Settings}
\label{sec:setting}

Our benchmark incorporates a total of 20 advanced MLLMs for systematic evaluation, including LLaVA-Next-Vicuna-7B \cite{liu2023improved}, LLaVA-Next-Mistral-7B \cite{liu2023improved}, LLaVA-Next-Llama3-8B \cite{li2024llavanext-strong}, LLaVA-OV-7B \cite{li2024llava}, Qwen2-VL-7B \cite{Qwen2-VL}, Qwen2.5-VL-7B \cite{Qwen2.5-VL}, InternVL3-8B \cite{zhu2025internvl3}, InternVL3.5-8B \cite{wang2025internvl3_5}, LLaVA-Next-34B \cite{liu2023improved}, Qwen2.5-VL-32B \cite{Qwen2.5-VL}, InternVL3-38B \cite{zhu2025internvl3}, InternVL3.5-38B \cite{wang2025internvl3_5}, Qwen2-VL-72B \cite{Qwen2-VL}, Qwen2.5-VL-72B \cite{Qwen2.5-VL}, QVQ-72B \cite{qvq-72b-preview}, InternVL3-78B \cite{zhu2025internvl3}, GPT-4o \cite{openaigpt4o}, GPT-5 \cite{openaigpt5}, Gemini-2.5-Flash \cite{gemini25flash}, and Gemini-2.5-Pro \cite{gemini25pro}. To ensure a fair and impartial performance assessment, we download the weights of open-source models from the Hugging Face platform\footnote{https://huggingface.co} and perform inference using the Hugging Face Transformers library\footnote{https://huggingface.co/docs/transformers/index}. For closed-source models, we employ the officially provided APIs for inference. Across all evaluated models, the temperature is fixed to 0 to reduce stochastic variation, and open-source models are evaluated under float32 precision. Experimental consistency is maintained wherever possible to minimize the influence of implementation differences.

\subsection{Performance Analysis}
\label{sec:analysis}

\begin{figure*}[t]
    \centering
    \includegraphics[width=\linewidth]{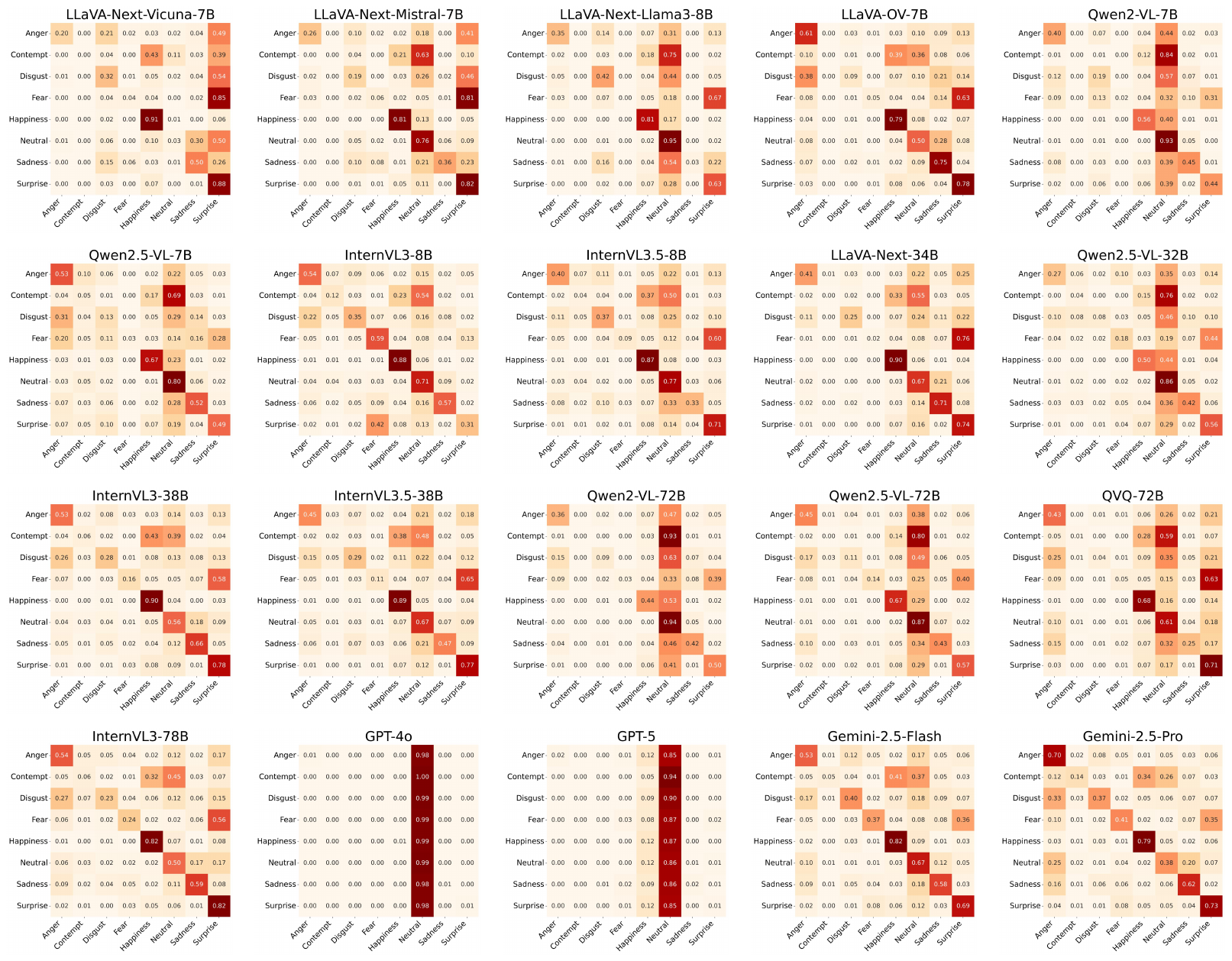}
    \caption{The confusion matrices of 20 evaluated MLLMs across various emotion categories on \benchmark{}.}
    \label{fig:cm}
    %\vspace{-2mm}
\end{figure*}

We provide the leaderboard on overall accuracy, with results shown in Fig. \ref{fig:bench}. The more comprehensive versions of task-level and category-level comparison results can be found in Table \ref{tab:main_cmp} and Table \ref{tab:cate_cmp}, respectively. Fig. \ref{fig:cm} showcases the confusion matrix results of 20 evaluated MLLMs.
Taking a close look at the performance of leading general-purpose MLLMs, we derive the following key observations:
\begin{enumerate}[leftmargin=*]
    \item[\ding{182}] \textbf{Off-the-shelf MLLMs demonstrate basic competence in recognizing emotions from facial images.} As shown in the leaderboard in Fig. \ref{fig:bench}, all evaluated MLLMs surpass the baseline of random guessing ($12.5\%$–$14.3\%$), indicating their preliminary capability in FER. Among them, four models, Gemini-2.5-Flash \cite{gemini25flash}, LLaVA-Next-34B \cite{liu2023improved}, InternVL3-38B \cite{zhu2025internvl3}, and InternVL3-8B \cite{zhu2025internvl3}, achieve an overall accuracy exceeding $60\%$. In addition, an anomalous phenomenon is observed: while Google’s closed-source Gemini-2.5-Flash \cite{gemini25flash} and Gemini-2.5-Pro \cite{gemini25pro} perform relatively well, OpenAI’s GPT-4o \cite{openaigpt4o} and GPT-5 \cite{openaigpt5} both score below $25\%$. Delving into their responses, we find that a majority of errors occur because these models fail to extract sufficient visual signals from blurry facial images for accurate judgment. Consequently, they tend to default to a "neutral" prediction for low-quality images (as shown in the last row of Fig. \ref{fig:cm}), highlighting their limitations in visual perception capabilities. Furthermore, models such as LLaVA-Next-Vicuna-7B \cite{liu2023improved}, LLaVA-OV-7B \cite{li2024llava}, and QVQ-72B \cite{qvq-72b-preview} demonstrate poor instruction-following capabilities. Consequently, extracting answers from specified <answer></answer> tags proves ineffective, requiring more complex format matching to obtain final predictions. In Table \ref{tab:main_cmp}, we make a fine-grained comparison of different datasets, and the results reveal variations in dataset difficulty. On simpler datasets like RAFDB \cite{li2017reliable,li2019reliable} and FERPlus \cite{BarsoumICMI2016}, some models achieve accuracy rates above $70\%$. In contrast, on more challenging datasets such as AffectNet \cite{mollahosseini2017affectnet} and SFEW2.0 \cite{zhang2024generalizable}, even the top-performing models reach only around $50\%$ accuracy. Further performance comparisons across different emotion categories in Table \ref{tab:cate_cmp} show that model capability varies by emotion. This aligns with natural intuition, as the frequency of different emotions in the real world leads to an uneven distribution in training data. For example, common emotions such as "happiness" are recognized with over $80\%$ accuracy by most MLLMs, whereas rare emotions like "contempt" lead to almost universal prediction failures. Notably, the powerful closed-source model Gemini-2.5-Pro \cite{gemini25pro} stands as an exception, achieving an accuracy of $20.49\%$.
    \item[\ding{183}] \textbf{The reasoning capability of general-purpose MLLMs remains a bottleneck for achieving interpretable and user-friendly FER.} Although current MLLMs have made remarkable progress in general reasoning tasks, most models have not yet demonstrated reasoning capabilities tailored for FER tasks due to limitations in training data. Most of the evaluated MLLMs tend to output emotion category predictions while neglecting the intermediate reasoning process based on facial clues. Although some models show preliminary reasoning attempts for FER tasks, such as QVQ-72B \cite{qvq-72b-preview}, which is built upon Qwen2-VL-72B \cite{Qwen2-VL} and further post-trained to enhance reasoning ability, their performance remains far from satisfactory. Compared to the unmodified Qwen2-VL-72B \cite{Qwen2-VL}, its accuracy even decreases from $48.19\%$ to $47.18\%$ on our \benchmark{}. This indicates that interpretable and user-friendly FER is still out of reach, underscoring the urgent need for a specialized FER foundation model capable of both high-quality reasoning and accurate recognition.
    \item[\ding{184}] \textbf{The emotional intelligence of general-purpose MLLMs is still limited and falls short of satisfactory performance on FER tasks.} As shown in the leaderboard in Fig. \ref{fig:bench}, even the best-performing closed-source model, Gemini-2.5-Flash \cite{gemini25flash}, achieves only an overall accuracy of $61.55\%$ on \benchmark{}, leaving substantial room for improvement. In Fig. \ref{fig:cm}, we visualize the confusion matrices of 20 MLLMs over 11K test samples. The horizontal axis represents the predicted emotions, while the vertical axis denotes the ground-truth emotion labels. Thus, each value in the $i$-th row and the $j$-th column indicates the proportion of samples with true emotion $i$ that were predicted as $j$ by the model. The diagonal entries correspond to the recall rates of each emotion category. We observe that for distinctly negative emotions such as "contempt", "disgust", and "fear", most MLLMs exhibit notably poor prediction accuracy. Collectively, these findings suggest that the emotional intelligence of off-the-shelf MLLMs, particularly their understanding of facial images, remains in its early stages and requires significant enhancement.
\end{enumerate}

\section{FER Foundation Model: \method{}-7B}

The above experimental results and analyses highlight the challenges faced by FER in the era of MLLMs. In this section, we seek to address these limitations through FER-aware post-training techniques. In Sec. \ref{sec:posttraining_data}, we introduce two carefully curated datasets, \method{}-RLVR-360K and \method{}-CoT-230K, which are subsequently utilized in conjunction with the post-training scheme detailed in Sec. \ref{sec:posttraining} to develop a specialized FER foundation model, termed \method{}-7B.

\begin{figure*}[t]
    \centering
    \includegraphics[width=\linewidth]{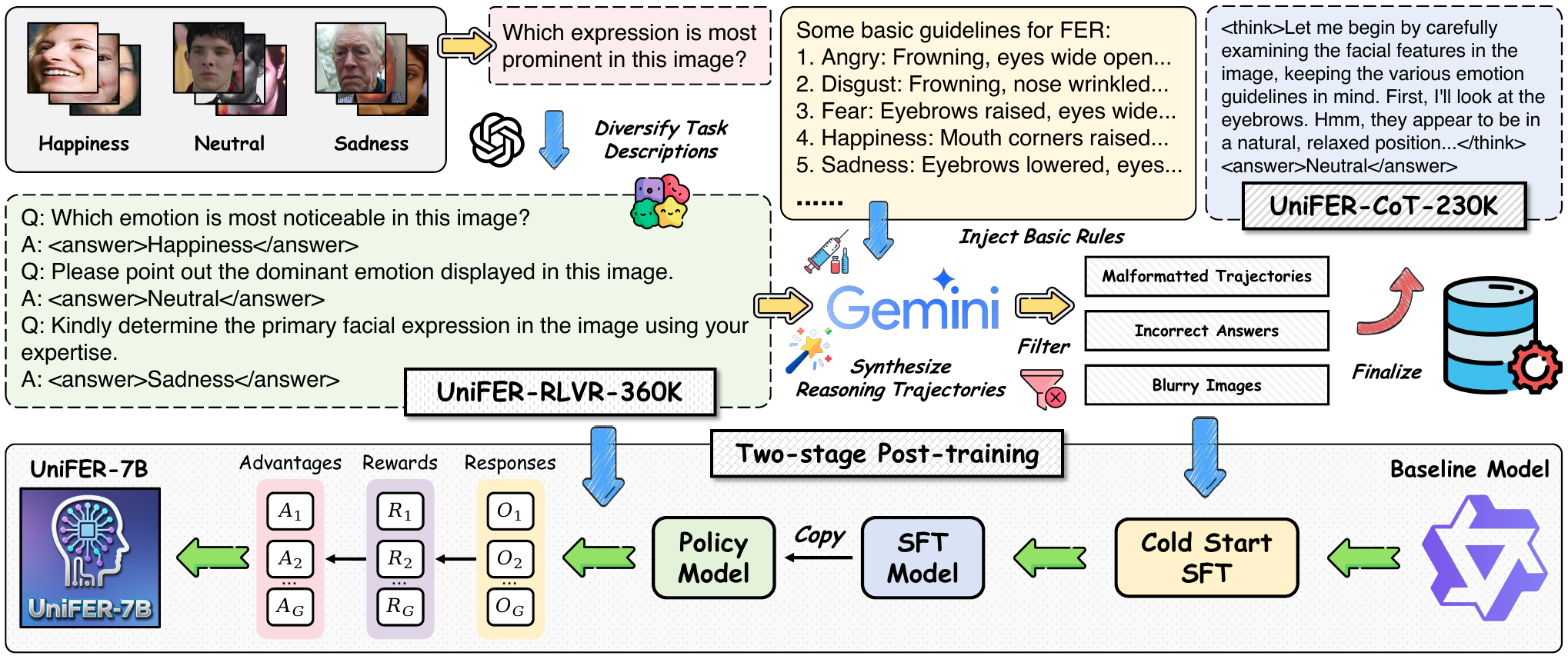}
    \caption{An overview of data curation and post-training pipeline. We curate two large-scale and high-quality datasets, and employ them for two-stage post-training, resulting in a unified and interpretable FER foundation model, \method{}-7B.}
    \label{fig:framework}
    %\vspace{-4mm}
\end{figure*}

\begin{table*}[ht]
\centering
\begin{minipage}{0.38\textwidth}
    \centering
    \caption{Key statistics of \method{}-RLVR-360K.}
    \label{tab:stat_unifer_rlvr_360k}
    \resizebox{\linewidth}{!}{%
    \begin{tabular}{l@{\hspace{0mm}}r}
    \toprule[1.2pt]
    \textbf{Statistic} & \textbf{Number} \\
    \midrule
    \textbf{Total Samples} & \bf 359,189 \\
    ~- Anger & 33,765 (9.4\%) \\
    ~- Contempt & 11,750 (3.3\%) \\
    ~- Disgust & 12,572 (3.5\%) \\
    ~- Fear & 14,737 (4.1\%) \\
    ~- Happiness & 147,370 (41.0\%) \\
    ~- Neutral & 87,920 (24.5\%) \\
    ~- Sadness & 35,601 (9.9\%) \\
    ~- Surprise & 15,474 (4.3\%) \\
    \midrule
    \textbf{Question} &  \\
    ~- Total Question Length & 23,933,781 \\
    ~- Maximum Question Length & 74 \\
    ~- Minimum Question Length & 58 \\
    ~- Average Question Length & 66.6\\
    \midrule
    \textbf{Answer} &  \\
    ~- Total Answer Length & 2,741,495 \\
    ~- Maximum Answer Length & 9 \\
    ~- Minimum Answer Length & 7 \\
    ~- Average Answer Length & 7.6 \\
    \bottomrule[1.2pt]
    \end{tabular}%
    }
\end{minipage}\hfill
\begin{minipage}{0.37\textwidth}
    \centering
    \caption{Key statistics of \method{}-CoT-230K.}
    \label{tab:stat_unifer_cot_230k}
    \resizebox{\linewidth}{!}{%
    \begin{tabular}{l@{\hspace{0mm}}r}
    \toprule[1.2pt]
    \textbf{Statistic} & \textbf{Number} \\
    \midrule
    \textbf{Total Samples} & \bf 229,394 \\
    ~- Anger & 21,123 (9.2\%) \\
    ~- Contempt & 6,983 (3.0\%) \\
    ~- Disgust & 10,877 (4.7\%) \\
    ~- Fear & 12,678 (5.5\%) \\
    ~- Happiness & 93,003 (40.5\%) \\
    ~- Neutral & 50,651 (22.1\%) \\
    ~- Sadness & 23,963 (10.4\%) \\
    ~- Surprise & 10,116 (4.4\%) \\
    \midrule
    \textbf{Question} &  \\
    ~- Total Question Length & 15,278,153 \\
    ~- Maximum Question Length & 74 \\
    ~- Minimum Question Length & 58 \\
    ~- Average Question Length & 66.6\\
    \midrule
    \textbf{Answer} &  \\
    ~- Total Answer Length & 100,085,757 \\
    ~- Maximum Answer Length & 4288 \\
    ~- Minimum Answer Length & 161 \\
    ~- Average Answer Length & 436.3 \\
    \bottomrule[1.2pt]
    \end{tabular}%
    }
\end{minipage}\hfill
\begin{minipage}{0.23\textwidth}
    \centering
    \includegraphics[width=\linewidth]{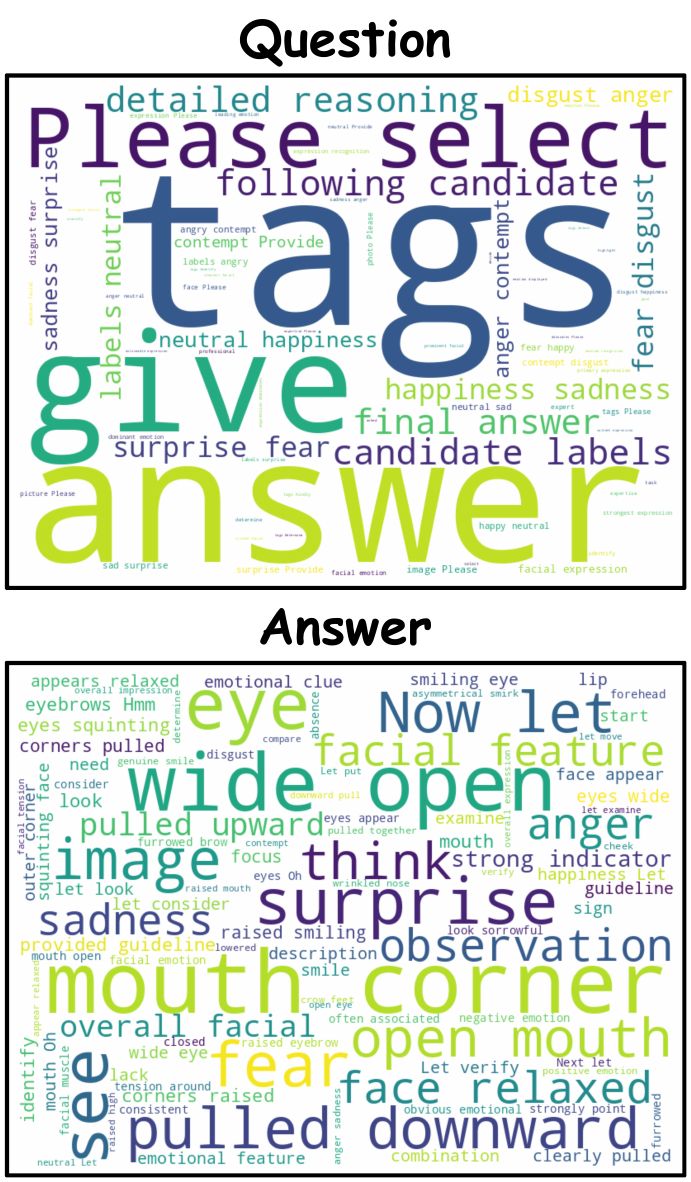}
    \vspace{-3mm}
    \captionsetup{type=figure}
    \caption{The word cloud visualization of questions (up) and answers (down) within \method{}-CoT-230K.}
    \label{fig:wordcloud}
\end{minipage}
\end{table*}

\subsection{Two High-quality Curated Datasets}
\label{sec:posttraining_data}

The success of post-training has been demonstrated across various domains and is regarded as a highly promising approach to enhancing the reasoning capabilities of both LLMs \cite{guo2025deepseek} and MLLMs \cite{feng2025video}. Motivated by this, we adopt a two-stage post-training strategy to improve the emotional intelligence of MLLMs tailored for FER tasks. Specifically, we first employ supervised fine-tuning (SFT) as a cold-start phase to teach the model how to reason following specific templates. Given that FER tasks provide explicit ground-truth emotion labels, we further leverage reinforcement learning with verifiable rewards (RLVR) in the second stage of post-training to enhance the model’s exploration ability. Consequently, two corresponding datasets are required for these stages. Interestingly, the dataset collection process proceeds in the reverse order of post-training: we first curate the \method{}-RLVR-360K dataset for RLVR, and then synthesize and filter reasoning trajectories based on it to construct the \method{}-CoT-230K dataset used for cold-start training.
In Fig. \ref{fig:framework}, we showcase an overview of the data curation process of \method{}-RLVR-360K and \method{}-CoT-230K, along with our two-stage post-training pipeline.

\subsubsection{\method{}-RLVR-360K}
As illustrated in the upper-left part of Fig. \ref{fig:framework}, we first collect facial images and their corresponding emotion labels from publicly available data, resulting in a total of 359,189 instances. We then convert the conventional image–label pairs into VQA samples by constructing a hand-crafted question example and associating it with emotion labels. To mitigate overfitting to fixed linguistic patterns during the post-training stage, we employ GPT-4o \cite{openaigpt4o} as a rewriting model to diversify the questions. Specifically, for the original question $\bm{q}^{(h)}$, we generate $K=100$ semantically equivalent but syntactically diverse variants, thereby enhancing the model’s linguistic robustness and generalization capability. Additionally, we enclose the answers within <answer></answer> tags to facilitate efficient result extraction during both RLVR training and evaluation phases. This process can be formulated as:
\begin{equation}
\mathcal{Q} = \{{\bm{q}^{(k)} \mid \bm{q}^{(k)} = \text{Rewrite}(\bm{q}^{(h)}; \text{GPT-4o})\}_{k=1}^{K}},
\end{equation}
\begin{equation}
    \mathcal{D}^{RLVR}=\{(\bm{x}_i, \bm{q}_i, \bm{a}_i) \mid \bm{q}_i \in \mathcal{Q} \}_{i=1}^{N},
\end{equation}
where $\bm{x}_i$, $\bm{q}_i$ and $\bm{a}_i$ denote the $i$-th facial image, question and answer, respectively.
$N=359,189$ is the number of samples in the \method{}-RLVR-360K dataset.
It should be noted that $\bm{q}_i$ is randomly selected from $\mathcal{Q}$.
The key statistics of \method{}-RLVR-360K, including emotion category distribution and QA length distribution, are presented in Table \ref{tab:stat_unifer_rlvr_360k}.

\subsubsection{\method{}-CoT-230K}

To equip the model with initial FER reasoning capabilities and facilitate efficient rollouts during the RLVR stage, we present a meticulously constructed CoT dataset containing high-quality FER reasoning trajectories for cold-start SFT training. Building upon the previously curated \method{}-RLVR-360K, we synthesize reasoning trajectories using the powerful closed-source MLLM Gemini-2.5-Flash \cite{gemini25flash}. The main challenge in this process lies in generating high-quality trajectories that exhibit long and coherent chains of reasoning.
To this end, we employ a two-stage strategy for data curation.

\vspace{1mm}
\noindent\textbf{Stage 1: Trajectory Synthesis through Backward Reasoning.} As shown in the upper-right part of Fig. \ref{fig:framework}, we first provide the trajectory generation model with paired facial images and corresponding ground-truth emotion labels, enabling it to reason backward from the answer to reconstruct the underlying reasoning process. To enhance the richness and consistency of the generated reasoning details, we inject some fundamental FER-specific rules into the model and guide it to perform multi-step, fine-grained reasoning. For example, the facial clues typically associated with the emotion "anger" include frowning, wide-open eyes, and mouth corners pulled downward. The generated trajectories and answers are also enclosed within <think></think> and <answer></answer> tags, respectively.
This process can be formulated as:
\begin{equation}
    \bm{t}_i = \text{Synthesize}(\bm{x}_i; \bm{a}_i; \bm{r}_i; \text{Gemini-2.5-Flash}),
\end{equation}
\begin{equation}
    \mathcal{D}^{SYN} =\{(\bm{x}_i, \bm{q}_i, \bm{t}_i, \bm{a}_i)\mid \bm{q}_i \in \mathcal{Q}\}_{i=1}^{N},
\end{equation}
where $\bm{r}_i$ and $\bm{t}_i$ refer to the basic rule for the $i$-th instance and its generated trajectory, respectively.

\vspace{1mm}
\noindent\textbf{Stage 2: Quality Control by Filtering Out Low-quality Instances.} 
To further improve the quality of our CoT dataset, we decide to filter out low-quality samples. We observe that the overall data quality is primarily affected by hallucinations from the trajectory generation model and the resolution of facial images. Accordingly, we implement a series of quality control strategies targeting these issues. Specifically, we filter out three types of samples:
\begin{enumerate}
    \item \textbf{Malformatted Trajectories.} We inspect all synthesized reasoning trajectories and remove those whose starting or ending positions do not contain the required <think></think> tags, as this indicates that the trajectory generation model hallucinated and failed to follow the instructed reasoning format.
    \item \textbf{Incorrect Answers.} We verify the final answers of the synthesized reasoning processes. If the generated answer is not enclosed within <answer></answer> tags or does not match the provided ground-truth emotion label, we discard the associated sample.
    \item \textbf{Blurry Images.} During the reasoning process, if any description suggests that the facial image is too blurry for the model to extract meaningful visual clues, we exclude the corresponding sample.
\end{enumerate}
The quality control process can be formulated as:
\begin{equation}
    \bm{g}_i = (\bm{x}_i, \bm{q}_i, \bm{t}_i, \bm{a}_i), 
\end{equation}
\begin{equation}
    \mathcal{D}^{CoT} = \{\bm{g}_i \mid \bm{g}_i \in \mathcal{D}^{SYN} \cap \bm{g}_i \notin \mathcal{D}^{Low} \}_{i=1}^{L}, 
\end{equation}
where $\mathcal{D}^{Low}$ denotes the set of low-quality samples defined above, and $L=229{,}394$ represents the total number of remaining instances in the \method{}-CoT-230K dataset.
Key dataset statistics and the word cloud visualization of \method{}-CoT-230K are presented in Table~\ref{tab:stat_unifer_cot_230k} and Fig.~\ref{fig:wordcloud}, respectively.

% \begin{figure}[t]
%     \centering
%     \includegraphics[width=\linewidth]{figs/wordcloud.pdf}
%     \caption{The word cloud visualization of questions and answers in \method{}-CoT-230K.}
%     \label{fig:wordcloud}
%     %\vspace{-4mm}
% \end{figure}

\subsection{Two-stage Post-training Scheme}
\label{sec:posttraining}

As shown in the bottom part of Fig. \ref{fig:framework}, we employ Qwen2.5-VL-7B \cite{Qwen2.5-VL} as the baseline model and perform two-stage post-training on the two curated datasets, resulting in a specialized FER foundation model, \method{}-7B.

\vspace{1mm}
\noindent\textbf{Cold Start Initialization.}
During the first stage, our goal is to teach the model to follow a pre-defined thinking format and reasoning path until it reaches the final prediction. To achieve this, we use SFT as a cold-start approach. For each trajectory and answer pair in $\mathcal{D}^{CoT}$, we force the model to predict the $j$-th reasoning step based on the given facial image $\bm{x}_i$, question $\bm{q}_i$, and all previous reasoning steps. In this formulation, SFT maximizes the log-likelihood of the target reasoning step $\bm{p}_{i}^{(j)}$:
\begin{equation}
    \max_{\theta} \sum_{i=1}^{L} \sum_{j=1}^{L_i} \log P_\theta \left( \bm{p}_{i}^{(j)} \mid \bm{x}_i, \bm{q}_i, \bm{p}^{(< j)}_{i} \right),
\end{equation}
where $\bm{p}_{i} = (\bm{t}_{i},\bm{a}_{i})$, $L_i$ is the length of $\bm{p}_{i}$, and $\theta$ denotes the model parameters.
After the cold-start SFT stage, the model learns to reason incrementally from visual clues, gradually arriving at the emotion prediction for a given facial image.

\vspace{1mm}
\noindent\textbf{Reinforcement Learning with Verified Rewards.}
After the cold-start initialization, we further employ Group-Relative Policy Optimization (GRPO) \cite{guo2025deepseek}, a ranking-based RLVR algorithm, to improve the exploration and reasoning capabilities of the SFT model for FER tasks.
Specifically, we adopt the SFT model as our policy model $\bm{\pi}_{\theta}$, enabling it to generate a set of $G$ responses $\mathcal{G} = \{ O_1,\cdots, O_G \}$, with each response $O_i$ assigned a rule-based reward $R_i$.
The reward is defined as:
\begin{equation}
    R = R^{\text{acc}} + R^{\text{format}},
\end{equation}
where $R^{\text{acc}}=1$ if the predicted facial expression matches the ground truth answer, otherwise $R^{\text{acc}}=0$; and $R^{\text{format}}=1$ if the response is enclosed within <think></think> and <answer></answer> tags, otherwise $R^{\text{format}}=0$.
Then, we can calculate the group-relative advantage $A_i$ as:
\begin{equation}
    A_i = \frac{R_i - \operatorname{mean}\left(\left\{R_j\right\}\right)}{\operatorname{std}\left(\left\{R_j\right\}\right)},
\end{equation}
where $\operatorname{mean}\left(\left\{R_j\right\}\right)$ and ${\operatorname{std}\left(\left\{R_j\right\}\right)}$ denote the mean and standard deviation of rewards within a group.
The policy model is updated using the following GRPO objective:
\begin{equation}
\begin{aligned}
&\mathcal{L}_{\mathrm{GRPO}}(\theta) = \mathbb{E}_{O_i \in \mathcal{G}} \left[ \frac{1}{G} \sum_{i=1}^{G} \min \left( \rho_{i} A_i, \right. \right. \\&
 \left. \left. \operatorname{clip} \left( \rho_{i}, 1-\epsilon, 1+\epsilon \right) A_i \right) \right] -  \beta D_{\mathrm{KL}} \left( \pi_{\theta} \| \pi_{\theta_{\mathrm{old}}} \right),
\end{aligned}
\end{equation}
where $\rho_{i} = \frac{\bm{\pi}_{\theta}\left(O_i \mid \bm{q}\right)}{\bm{\pi}_{\theta_{\mathrm{old}}}\left(O_i \mid \bm{q}\right)}$ represents the importance sampling ratio, $\epsilon$ is the clipping parameter that bounds the probability ratio updates, and $D_{\mathrm{KL}}$ quantifies the Kullback-Leibler (KL) divergence between the current policy $\bm{\pi}_{\theta}$ and its predecessor $\bm{\pi}_{\theta_{\mathrm{old}}}$.
The hyperparameter $\beta$ governs the magnitude of the KL constraint. 
This formulation ensures training stability by preventing excessive policy updates, while simultaneously favoring actions that yield superior relative advantages.
After RLVR training, we obtain a specialized FER foundation model named \method{}-7B, possessing powerful capabilities for facial expression reasoning and recognition.

\section{Further Experiments}

In this section, we make additional experiments to demonstrate the advantages of our \method{}-7B. Sec. \ref{sec:cmp_unifer7b} showcases the comparison among the baseline model, previous SOTA, and \method{}-7B. Sec. \ref{sec:ablation_unifer7b} provides an ablation study for our two-stage post-training scheme. Sec. \ref{sec:case_study_unifer7b} presents a case study of \method{}-7B and the competing approaches.

\begin{figure*}[t]
    \centering
    \includegraphics[width=\linewidth]{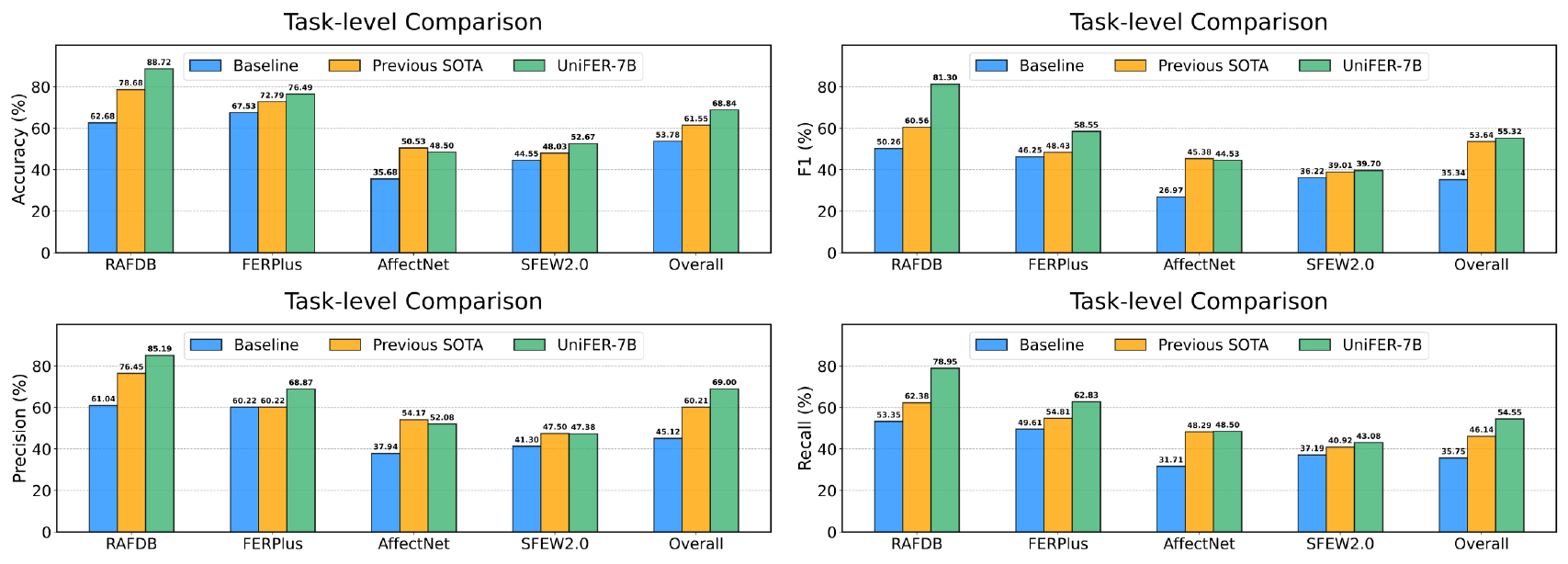}
    \caption{Task-level comparison (in $\%$) across the baseline model, previous SOTA, and our \method{}-7B.}
    \label{fig:task_cmp}
    %\vspace{-3mm}
\end{figure*}
\begin{figure*}[ht]
    \centering
    \includegraphics[width=\linewidth]{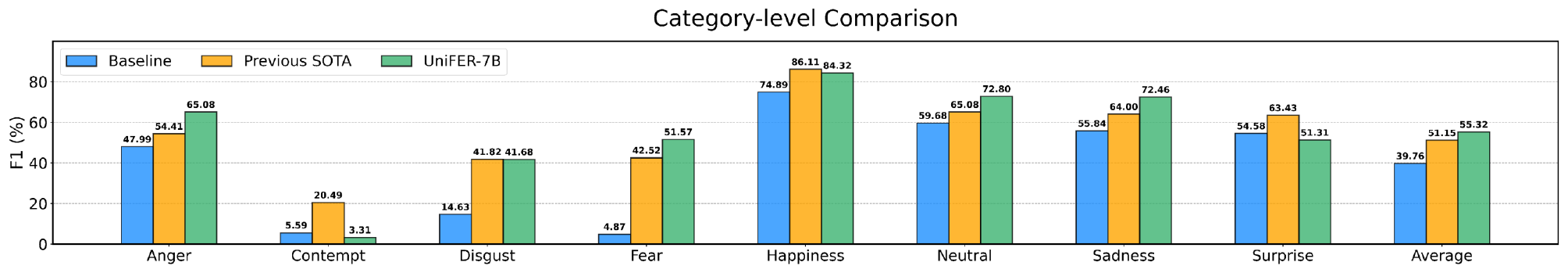}
    \caption{Category-level comparison (in $\%$) across the baseline model, previous SOTA, and our \method{}-7B.}
    \label{fig:cat_cmp}
    %\vspace{-4mm}
\end{figure*}

\begin{table*}[ht]
\centering
\caption{Ablation study (in $\%$) on \benchmark{}. Best results are marked in \textbf{bold}.}
\label{tab:task_ablation}
\resizebox{\linewidth}{!}{%
% \begin{tabular}{lcccccccccc}
\begin{tabular}{l@{\hspace{5mm}}c@{\hspace{5mm}}c@{\hspace{5mm}}c@{\hspace{5mm}}c@{\hspace{5mm}}c@{\hspace{5mm}}c@{\hspace{5mm}}c@{\hspace{5mm}}c@{\hspace{5mm}}c@{\hspace{5mm}}c}
\toprule[1.2pt]
\multirow{2.5}{*}{\textbf{Model}} & \multicolumn{2}{c}{~\hspace{-7mm}\textbf{RAFDB}} & \multicolumn{2}{c}{~\hspace{-7mm}\textbf{FERPlus}} & \multicolumn{2}{c}{~\hspace{-7mm}\textbf{AffectNet}} & \multicolumn{2}{c}{~\hspace{-7mm}\textbf{SFEW2.0}} & \multicolumn{2}{c}{~\hspace{-5mm}\textbf{Overall}} \\
 \cmidrule{2-11}& \textbf{Acc} & \textbf{F1} & \textbf{Acc} & \textbf{F1} & \textbf{Acc} & \textbf{F1} & \textbf{Acc} & \textbf{F1} & \textbf{Acc} & \textbf{F1} \\
\midrule
Qwen2.5-VL-7B \cite{Qwen2.5-VL} & 62.68 & 50.26 & 67.53 & 46.25 & 35.68 & 26.97 & 44.55 & 36.22 & 53.78 & 35.34 \\
\quad $+$ Cold Start & 86.83 &	78.64 &	74.14 &	55.31 &	47.18 &	42.15 &	51.51 &	37.89 &	67.03 &	52.67 \\
\quad $+$ Cold Start$\&$RLVR & \textbf{88.72} &	\textbf{81.30} &	\textbf{76.49} &	\textbf{58.55} &	\textbf{48.50} &	\textbf{44.53} &	\textbf{52.67} &	\textbf{39.70} &	\textbf{68.84} &	\textbf{55.32}  \\
\bottomrule[1.2pt]
\end{tabular}%
}
\end{table*}

% \begin{table*}[ht]
% \centering
% \caption{Category-level ablation study (in $\%$) on \benchmark{}. Best results are marked in \textbf{bold}.}
% \label{tab:cate_ablation}
% \resizebox{\linewidth}{!}{%
% \begin{tabular}{lccccccccc}
% \toprule[1.2pt]
% \textbf{Model} & \textbf{Anger} & \textbf{Contempt} & \textbf{Disgust} & \textbf{Fear} & \textbf{Happiness} & \textbf{Neutral} & \textbf{Sadness} & \textbf{Surprise} & \textbf{Avg} \\
% \midrule
% Qwen2.5-VL-7B & 47.99 & \textbf{5.59} & 14.63 & 4.87 & 74.89 & 59.68 & 55.84 & 54.58 & \textbf{39.76} \\
% \quad $+$ Cold Start & 62.66 &	2.96 &	35.63 &	46.86 &	82.30 &	\textbf{73.92} &	69.60 &	47.44 	&52.67  \\
% \quad $+$ Cold Start$\&$RLVR & \textbf{65.08} &	3.31 & \textbf{41.68} &	\textbf{51.57} &	\textbf{84.32} &	72.80 &	\textbf{72.46} &	51.31 	& \textbf{55.32} \\
% \bottomrule[1.2pt]
% \end{tabular}%
% }
% \end{table*}

\subsection{Comparison with Baseline and Previous SOTA}
\label{sec:cmp_unifer7b}

Fig. \ref{fig:task_cmp} presents the task-level comparison among the baseline model (Qwen2.5-VL-7B \cite{Qwen2.5-VL}), previous SOTA, and our \method{}-7B on \benchmark{}.
As shown, compared with the baseline, \method{}-7B achieves significant improvements across all four evaluation metrics on each subset of \benchmark{} (RAFDB, FEPlus, AffectNet, and SFEW2.0) as well as in the overall setting.
When compared to the previous SOTA, \method{}-7B surpasses it in the vast majority of scenarios and metrics.
Notably, the highest score on the \benchmark{} leaderboard was previously held by Gemini-2.5-Flash \cite{gemini25flash} at $61.55\%$, whereas our \method{}-7B establishes a new record of $68.84\%$, marking a substantial improvement.
In the category-level comparison shown in Fig. \ref{fig:cat_cmp}, we present a fine-grained analysis across different emotion categories. We similarly observe that \method{}-7B demonstrates clear advantages over the baseline model in most (6 out of 8) emotion categories. Overall, \method{}-7B achieves an average F1 score of $55.32\%$, surpassing the baseline and previous SOTA by $15.56\%$ and $4.17\%$, respectively.
We attribute this improvement to the effectiveness of our two-stage post-training scheme, which enhances \method{}-7B’s ability to capture emotion-aware visual clues and derive precise reasoning processes, ultimately leading to more accurate emotion predictions.

\subsection{Ablation Study}
\label{sec:ablation_unifer7b}

To further validate the effectiveness of our two-stage post-training scheme, we conduct an ablation study, with the results presented in Table \ref{tab:task_ablation}.
It can be observed that after applying the cold-start initialization to the baseline model Qwen2.5-VL-7B \cite{Qwen2.5-VL}, its performance improves substantially across both metrics in the overall setting and all four subsets. Specifically, the cold-start SFT stage increases the overall accuracy and F1 score by $13.25\%$ and $17.33\%$, respectively, which represents a significant improvement. This demonstrates that SFT effectively equips the model with more detailed reasoning process and accurate prediction capability.
Building upon this, training with RLVR further enhances performance beyond the SFT stage, yielding consistent improvements across all evaluation settings. On top of the SFT model, RLVR brings an additional $1.81\%$ gain in accuracy and $2.65\%$ in F1 score, establishing a new SOTA result.
These findings indicate that RLVR further enhances the model’s capacity for exploration and reasoning, thereby leading to improved recognition performance. In summary, both stages of post-training are indispensable and jointly contribute to the exceptional improvement in FER performance.

\begin{figure*}[th]
    \centering
    \includegraphics[width=\linewidth]{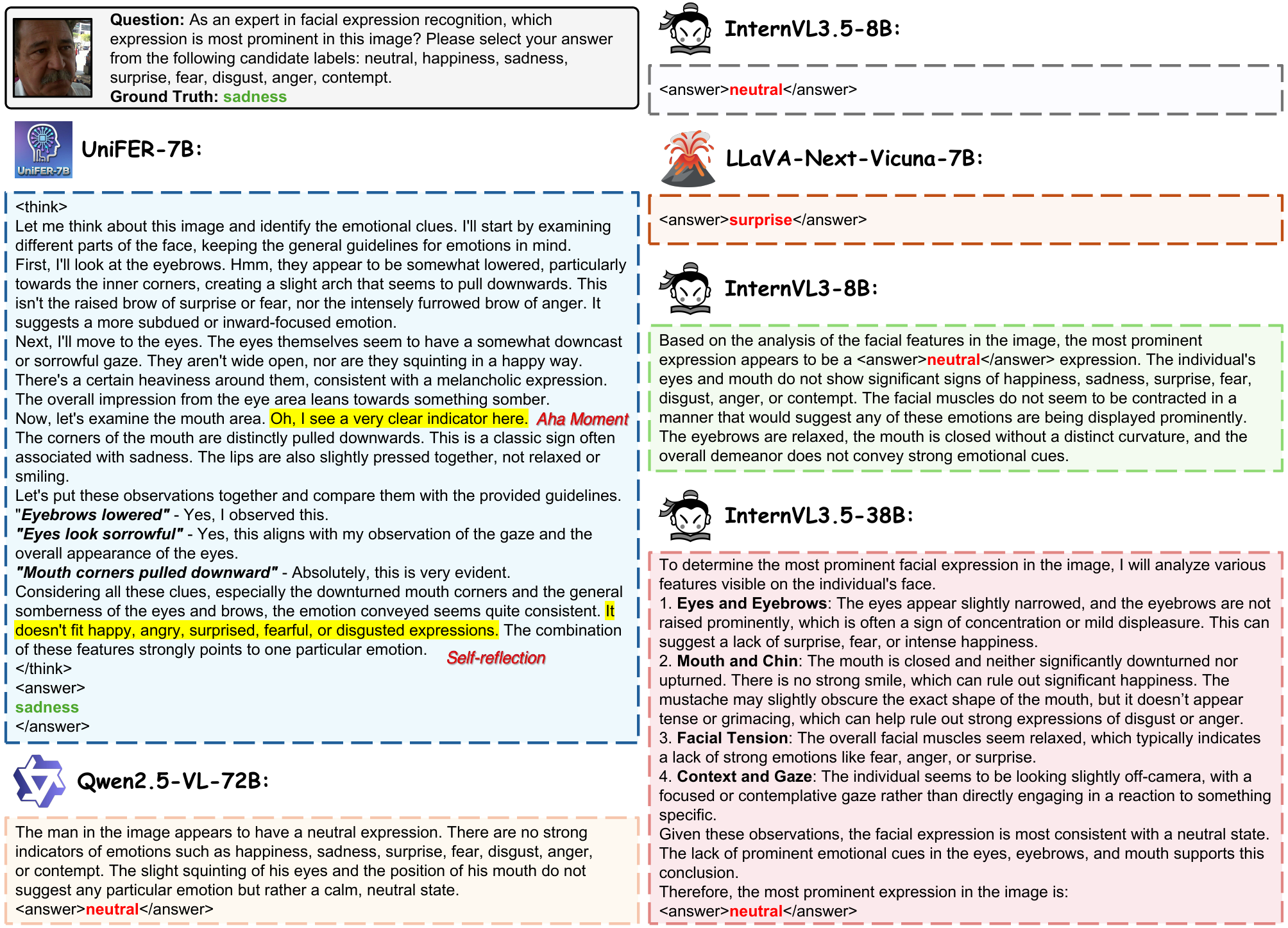}
    \caption{An illustrative example of a question and the responses generated by various MLLMs on \benchmark{}.}
    \label{fig:case}
    %\vspace{-4mm}
\end{figure*}

%\vspace{-3mm}
\subsection{Case Study}
\label{sec:case_study_unifer7b}

In Fig. \ref{fig:case}, we present an example question along with the corresponding responses generated by different MLLMs on \benchmark{}. It can be observed that some models (\emph{e.g.}, InternVL3.5-8B \cite{wang2025internvl3_5} and LLaVA-Next-Vicuna-7B \cite{liu2023improved}) incorrectly recognize the emotion and fail to provide any reasoning process, making it impossible to understand the rationale behind their predictions. Other models (\emph{e.g.}, Qwen2.5-VL-72B \cite{Qwen2.5-VL}, InternVL3-8B \cite{zhu2025internvl3}, and InternVL3.5-38B \cite{wang2025internvl3_5}) are capable of step-by-step reasoning based on facial clues, attending to key regions such as the eyes and mouth, yet they often generate factually inconsistent observations.
For instance, the response from InternVL3-8B \cite{zhu2025internvl3} describes the person in the image as having “relaxed eyebrows” and “a closed mouth without a distinct curvature”, which is clearly inconsistent with the visual evidence. Although InternVL3.5-38B \cite{wang2025internvl3_5} exhibits seemingly more structured reasoning by focusing on emotional clues related to eyes and eyebrows, mouth and chin, facial tension, and context and gaze, it still introduces logical errors during intermediate steps and lacks effective self-reflection, preventing it from synthesizing prior reasoning and ultimately leading to incorrect predictions.
In contrast, our \method{}-7B is the only model capable of producing both high-quality reasoning traces and accurate emotion recognition results. When analyzing facial clues, \method{}-7B begins by attending to the eyebrows, then gradually shifts attention to the eyes and mouth, comparing its observations against the injected emotional principles to reach a reliable conclusion. Remarkably, upon focusing on the mouth region, \method{}-7B correctly infers the emotion of sadness, explicitly noting, “Oh, I see a very clear indicator here”. This phenomenon marks a notable “aha moment” in the emergence of multimodal reasoning within the FER domain.
Toward the end of its reasoning, we further observe that \method{}-7B demonstrates self-reflective behavior by accurately revisiting its prior steps and excluding alternative emotional categories based on the accumulated evidence.
These findings collectively highlight the significant advancements that \method{}-7B brings to facial expression reasoning.

\section{Conclusion}

In this paper, we revisited the classic task of facial expression recognition (FER) in the era of multimodal large language models (MLLMs). We first introduced \benchmark{}, an open and fair benchmark designed to evaluate the emotional intelligence of cutting-edge MLLMs on FER tasks. Through a comprehensive analysis of the evaluation results, we identified a significant limitation in the emotion reasoning capability of existing MLLMs. To address this, we constructed two large-scale and high-quality post-training datasets, namely \method{}-CoT-230K and \method{}-RLVR-360K, and proposed a two-stage post-training scheme that combines cold-start supervised fine-tuning (SFT) with reinforcement learning with verifiable rewards (RLVR). Based on this framework, we successfully developed a unified and interpretable FER foundation model, termed \method{}-7B. Further experimental results demonstrated that \method{}-7B achieves outstanding performance in both facial expression reasoning and recognition, establishing a new SOTA for this field. In future works, we plan to extend multimodal reasoning techniques to broader areas of affective computing, such as video-based dynamic settings and omnimodal scenarios.

%\clearpage

\section*{Acknowledgement}
The authors are grateful to the anonymous reviewers for critically reading the manuscript and for giving important suggestions to improve their paper.

\bibliographystyle{IEEEtran}
\bibliography{rec}

\begin{thebibliography}{10}
\providecommand{\url}[1]{#1}
\csname url@samestyle\endcsname
\providecommand{\newblock}{\relax}
\providecommand{\bibinfo}[2]{#2}
\providecommand{\BIBentrySTDinterwordspacing}{\spaceskip=0pt\relax}
\providecommand{\BIBentryALTinterwordstretchfactor}{4}
\providecommand{\BIBentryALTinterwordspacing}{\spaceskip=\fontdimen2\font plus
\BIBentryALTinterwordstretchfactor\fontdimen3\font minus \fontdimen4\font\relax}
\providecommand{\BIBforeignlanguage}[2]{{%
\expandafter\ifx\csname l@#1\endcsname\relax
\typeout{** WARNING: IEEEtran.bst: No hyphenation pattern has been}%
\typeout{** loaded for the language `#1'. Using the pattern for}%
\typeout{** the default language instead.}%
\else
\language=\csname l@#1\endcsname
\fi
#2}}
\providecommand{\BIBdecl}{\relax}
\BIBdecl

\bibitem{li2020deep}
S.~Li and W.~Deng, ``Deep facial expression recognition: A survey,'' \emph{IEEE transactions on affective computing}, vol.~13, no.~3, pp. 1195--1215, 2020.

\bibitem{tian2011facial}
Y.~Tian, T.~Kanade, and J.~F. Cohn, ``Facial expression recognition,'' in \emph{Handbook of face recognition}.\hskip 1em plus 0.5em minus 0.4em\relax Springer, 2011, pp. 487--519.

\bibitem{kumari2015facial}
J.~Kumari, R.~Rajesh, and K.~Pooja, ``Facial expression recognition: A survey,'' \emph{Procedia computer science}, vol.~58, pp. 486--491, 2015.

\bibitem{mao2025facial}
S.~Mao, X.~Li, F.~Zhang, X.~Peng, and Y.~Yang, ``Facial action units as a joint dataset training bridge for facial expression recognition,'' \emph{IEEE Transactions on Multimedia}, 2025.

\bibitem{zhang2011facial}
L.~Zhang and D.~Tjondronegoro, ``Facial expression recognition using facial movement features,'' \emph{IEEE transactions on affective computing}, vol.~2, no.~4, pp. 219--229, 2011.

\bibitem{chattopadhyay2020facial}
J.~Chattopadhyay, S.~Kundu, A.~Chakraborty, and J.~S. Banerjee, ``Facial expression recognition for human computer interaction,'' in \emph{New Trends in Computational Vision and Bio-inspired Computing: Selected works presented at the ICCVBIC 2018, Coimbatore, India}.\hskip 1em plus 0.5em minus 0.4em\relax Springer, 2020, pp. 1181--1192.

\bibitem{cowie2001emotion}
R.~Cowie, E.~Douglas-Cowie, N.~Tsapatsoulis, G.~Votsis, S.~Kollias, W.~Fellenz, and J.~G. Taylor, ``Emotion recognition in human-computer interaction,'' \emph{IEEE Signal processing magazine}, vol.~18, no.~1, pp. 32--80, 2001.

\bibitem{kakarla2014real}
M.~Kakarla and G.~R.~M. Reddy, ``A real time facial emotion recognition using depth sensor and interfacing with second life based virtual 3d avatar,'' in \emph{International Conference on Recent Advances and Innovations in Engineering (ICRAIE-2014)}.\hskip 1em plus 0.5em minus 0.4em\relax IEEE, 2014, pp. 1--7.

\bibitem{yang2011facial}
S.~Yang and B.~Bhanu, ``Facial expression recognition using emotion avatar image,'' in \emph{2011 IEEE International Conference on Automatic Face \& Gesture Recognition (FG)}.\hskip 1em plus 0.5em minus 0.4em\relax IEEE, 2011, pp. 866--871.

\bibitem{tacconi2008activity}
D.~Tacconi, O.~Mayora, P.~Lukowicz, B.~Arnrich, C.~Setz, G.~Troster, and C.~Haring, ``Activity and emotion recognition to support early diagnosis of psychiatric diseases,'' in \emph{2008 second international conference on pervasive computing technologies for healthcare}.\hskip 1em plus 0.5em minus 0.4em\relax IEEE, 2008, pp. 100--102.

\bibitem{pepa2021automatic}
L.~Pepa, L.~Spalazzi, M.~Capecci, and M.~G. Ceravolo, ``Automatic emotion recognition in clinical scenario: a systematic review of methods,'' \emph{IEEE Transactions on Affective Computing}, vol.~14, no.~2, pp. 1675--1695, 2021.

\bibitem{simonyan2014very}
K.~Simonyan and A.~Zisserman, ``Very deep convolutional networks for large-scale image recognition,'' \emph{arXiv preprint arXiv:1409.1556}, 2014.

\bibitem{he2016deep}
K.~He, X.~Zhang, S.~Ren, and J.~Sun, ``Deep residual learning for image recognition,'' in \emph{Proceedings of the IEEE conference on computer vision and pattern recognition}, 2016, pp. 770--778.

\bibitem{dosovitskiy2020image}
A.~Dosovitskiy, L.~Beyer, A.~Kolesnikov, D.~Weissenborn, X.~Zhai, T.~Unterthiner, M.~Dehghani, M.~Minderer, G.~Heigold, S.~Gelly \emph{et~al.}, ``An image is worth 16x16 words: Transformers for image recognition at scale,'' \emph{arXiv preprint arXiv:2010.11929}, 2020.

\bibitem{liu2021swin}
Z.~Liu, Y.~Lin, Y.~Cao, H.~Hu, Y.~Wei, Z.~Zhang, S.~Lin, and B.~Guo, ``Swin transformer: Hierarchical vision transformer using shifted windows,'' in \emph{Proceedings of the IEEE/CVF international conference on computer vision}, 2021, pp. 10\,012--10\,022.

\bibitem{zhang2021relative}
Y.~Zhang, C.~Wang, and W.~Deng, ``Relative uncertainty learning for facial expression recognition,'' \emph{Advances in Neural Information Processing Systems}, vol.~34, pp. 17\,616--17\,627, 2021.

\bibitem{zeng2022face2exp}
D.~Zeng, Z.~Lin, X.~Yan, Y.~Liu, F.~Wang, and B.~Tang, ``Face2exp: Combating data biases for facial expression recognition,'' in \emph{Proceedings of the IEEE/CVF conference on computer vision and pattern recognition}, 2022, pp. 20\,291--20\,300.

\bibitem{zhang2022learn}
Y.~Zhang, C.~Wang, X.~Ling, and W.~Deng, ``Learn from all: Erasing attention consistency for noisy label facial expression recognition,'' in \emph{European Conference on Computer Vision}.\hskip 1em plus 0.5em minus 0.4em\relax Springer, 2022, pp. 418--434.

\bibitem{chen2021understanding}
Y.~Chen and J.~Joo, ``Understanding and mitigating annotation bias in facial expression recognition,'' in \emph{Proceedings of the IEEE/CVF International Conference on Computer Vision}, 2021, pp. 14\,980--14\,991.

\bibitem{lian2024gpt}
Z.~Lian, L.~Sun, H.~Sun, K.~Chen, Z.~Wen, H.~Gu, B.~Liu, and J.~Tao, ``Gpt-4v with emotion: A zero-shot benchmark for generalized emotion recognition,'' \emph{Information Fusion}, vol. 108, p. 102367, 2024.

\bibitem{feng2025video}
K.~Feng, K.~Gong, B.~Li, Z.~Guo, Y.~Wang, T.~Peng, J.~Wu, X.~Zhang, B.~Wang, and X.~Yue, ``Video-r1: Reinforcing video reasoning in mllms,'' \emph{arXiv preprint arXiv:2503.21776}, 2025.

\bibitem{li2017reliable}
S.~Li, W.~Deng, and J.~Du, ``Reliable crowdsourcing and deep locality-preserving learning for expression recognition in the wild,'' in \emph{2017 IEEE Conference on Computer Vision and Pattern Recognition (CVPR)}.\hskip 1em plus 0.5em minus 0.4em\relax IEEE, 2017, pp. 2584--2593.

\bibitem{li2019reliable}
S.~Li and W.~Deng, ``Reliable crowdsourcing and deep locality-preserving learning for unconstrained facial expression recognition,'' \emph{IEEE Transactions on Image Processing}, vol.~28, no.~1, pp. 356--370, 2019.

\bibitem{BarsoumICMI2016}
E.~Barsoum, C.~Zhang, C.~Canton~Ferrer, and Z.~Zhang, ``Training deep networks for facial expression recognition with crowd-sourced label distribution,'' in \emph{ACM International Conference on Multimodal Interaction (ICMI)}, 2016.

\bibitem{mollahosseini2017affectnet}
A.~Mollahosseini, B.~Hasani, and M.~H. Mahoor, ``Affectnet: A database for facial expression, valence, and arousal computing in the wild,'' \emph{IEEE Transactions on Affective Computing}, vol.~10, no.~1, pp. 18--31, 2017.

\bibitem{zhang2024generalizable}
Y.~Zhang, X.~Zheng, C.~Liang, J.~Hu, and W.~Deng, ``Generalizable facial expression recognition,'' in \emph{European Conference on Computer Vision}.\hskip 1em plus 0.5em minus 0.4em\relax Springer, 2024, pp. 231--248.

\bibitem{xu2025towards}
F.~Xu, Q.~Hao, Z.~Zong, J.~Wang, Y.~Zhang, J.~Wang, X.~Lan, J.~Gong, T.~Ouyang, F.~Meng \emph{et~al.}, ``Towards large reasoning models: A survey of reinforced reasoning with large language models,'' \emph{arXiv preprint arXiv:2501.09686}, 2025.

\bibitem{guo2025deepseek}
D.~Guo, D.~Yang, H.~Zhang, J.~Song, R.~Zhang, R.~Xu, Q.~Zhu, S.~Ma, P.~Wang, X.~Bi \emph{et~al.}, ``Deepseek-r1: Incentivizing reasoning capability in llms via reinforcement learning,'' \emph{arXiv preprint arXiv:2501.12948}, 2025.

\bibitem{patil2025advancing}
A.~Patil and A.~Jadon, ``Advancing reasoning in large language models: Promising methods and approaches,'' \emph{arXiv preprint arXiv:2502.03671}, 2025.

\bibitem{Qwen2.5-VL}
S.~Bai, K.~Chen, X.~Liu, J.~Wang, W.~Ge, S.~Song, K.~Dang, P.~Wang, S.~Wang, J.~Tang, H.~Zhong, Y.~Zhu, M.~Yang, Z.~Li, J.~Wan, P.~Wang, W.~Ding, Z.~Fu, Y.~Xu, J.~Ye, X.~Zhang, T.~Xie, Z.~Cheng, H.~Zhang, Z.~Yang, H.~Xu, and J.~Lin, ``Qwen2.5-vl technical report,'' \emph{arXiv preprint arXiv:2502.13923}, 2025.

\bibitem{zhu2025internvl3}
J.~Zhu, W.~Wang, Z.~Chen, Z.~Liu, S.~Ye, L.~Gu, H.~Tian, Y.~Duan, W.~Su, J.~Shao \emph{et~al.}, ``Internvl3: Exploring advanced training and test-time recipes for open-source multimodal models,'' \emph{arXiv preprint arXiv:2504.10479}, 2025.

\bibitem{openaigpt5}
\BIBentryALTinterwordspacing
OpenAI, ``Gpt-5 system card,'' 2025. [Online]. Available: \url{https://cdn.openai.com/gpt-5-system-card.pdf}
\BIBentrySTDinterwordspacing

\bibitem{gemini25pro}
\BIBentryALTinterwordspacing
Google, ``Gemini 2.5 pro preview model card,'' 2025. [Online]. Available: \url{https://storage.googleapis.com/model-cards/documents/gemini-2.5-pro-preview.pdf}
\BIBentrySTDinterwordspacing

\bibitem{revina2021survey}
I.~M. Revina and W.~S. Emmanuel, ``A survey on human face expression recognition techniques,'' \emph{Journal of King Saud University-Computer and Information Sciences}, vol.~33, no.~6, pp. 619--628, 2021.

\bibitem{bettadapura2012face}
V.~Bettadapura, ``Face expression recognition and analysis: the state of the art,'' \emph{arXiv preprint arXiv:1203.6722}, 2012.

\bibitem{huang2019facial}
Y.~Huang, F.~Chen, S.~Lv, and X.~Wang, ``Facial expression recognition: A survey,'' \emph{Symmetry}, vol.~11, no.~10, p. 1189, 2019.

\bibitem{ng2003sift}
P.~C. Ng and S.~Henikoff, ``Sift: Predicting amino acid changes that affect protein function,'' \emph{Nucleic acids research}, vol.~31, no.~13, pp. 3812--3814, 2003.

\bibitem{dalal2005histograms}
N.~Dalal and B.~Triggs, ``Histograms of oriented gradients for human detection,'' in \emph{2005 IEEE computer society conference on computer vision and pattern recognition (CVPR'05)}, vol.~1.\hskip 1em plus 0.5em minus 0.4em\relax Ieee, 2005, pp. 886--893.

\bibitem{shan2009facial}
C.~Shan, S.~Gong, and P.~W. McOwan, ``Facial expression recognition based on local binary patterns: A comprehensive study,'' \emph{Image and vision Computing}, vol.~27, no.~6, pp. 803--816, 2009.

\bibitem{liu2002gabor}
C.~Liu and H.~Wechsler, ``Gabor feature based classification using the enhanced fisher linear discriminant model for face recognition,'' \emph{IEEE Transactions on Image processing}, vol.~11, no.~4, pp. 467--476, 2002.

\bibitem{hu2008multi}
Y.~Hu, Z.~Zeng, L.~Yin, X.~Wei, X.~Zhou, and T.~S. Huang, ``Multi-view facial expression recognition,'' in \emph{2008 8th IEEE International Conference on Automatic Face \& Gesture Recognition}.\hskip 1em plus 0.5em minus 0.4em\relax IEEE, 2008, pp. 1--6.

\bibitem{luo2013facial}
Y.~Luo, C.-m. Wu, and Y.~Zhang, ``Facial expression recognition based on fusion feature of pca and lbp with svm,'' \emph{Optik-International Journal for Light and Electron Optics}, vol. 124, no.~17, pp. 2767--2770, 2013.

\bibitem{pietikainen2011computer}
M.~Pietik{\"a}inen, A.~Hadid, G.~Zhao, and T.~Ahonen, \emph{Computer vision using local binary patterns}.\hskip 1em plus 0.5em minus 0.4em\relax Springer Science \& Business Media, 2011, vol.~40.

\bibitem{cheng2023semi}
Z.~Cheng, Y.~Lin, Z.~Chen, X.~Li, S.~Mao, F.~Zhang, D.~Ding, B.~Zhang, and X.~Peng, ``Semi-supervised multimodal emotion recognition with expression mae,'' in \emph{Proceedings of the 31st ACM International Conference on Multimedia}, 2023, pp. 9436--9440.

\bibitem{she2021dive}
J.~She, Y.~Hu, H.~Shi, J.~Wang, Q.~Shen, and T.~Mei, ``Dive into ambiguity: Latent distribution mining and pairwise uncertainty estimation for facial expression recognition,'' in \emph{Proceedings of the IEEE/CVF conference on computer vision and pattern recognition}, 2021, pp. 6248--6257.

\bibitem{wang2020suppressing}
K.~Wang, X.~Peng, J.~Yang, S.~Lu, and Y.~Qiao, ``Suppressing uncertainties for large-scale facial expression recognition,'' in \emph{Proceedings of the IEEE/CVF conference on computer vision and pattern recognition}, 2020, pp. 6897--6906.

\bibitem{zhang2024leaf}
F.~Zhang, Z.-Q. Cheng, J.~Zhao, X.~Peng, and X.~Li, ``Leaf: unveiling two sides of the same coin in semi-supervised facial expression recognition,'' \emph{arXiv preprint arXiv:2404.15041}, 2024.

\bibitem{xue2021transfer}
F.~Xue, Q.~Wang, and G.~Guo, ``Transfer: Learning relation-aware facial expression representations with transformers,'' in \emph{Proceedings of the IEEE/CVF International Conference on Computer Vision}, 2021, pp. 3601--3610.

\bibitem{dai2023instructblip}
W.~Dai, J.~Li, D.~Li, A.~Tiong, J.~Zhao, W.~Wang, B.~Li, P.~N. Fung, and S.~Hoi, ``Instructblip: Towards general-purpose vision-language models with instruction tuning,'' \emph{Advances in neural information processing systems}, vol.~36, pp. 49\,250--49\,267, 2023.

\bibitem{liu2023visual}
H.~Liu, C.~Li, Q.~Wu, and Y.~J. Lee, ``Visual instruction tuning,'' \emph{Advances in neural information processing systems}, vol.~36, pp. 34\,892--34\,916, 2023.

\bibitem{huang2025visual}
J.~Huang, J.~Zhang, K.~Jiang, H.~Qiu, X.~Zhang, L.~Shao, S.~Lu, and D.~Tao, ``Visual instruction tuning towards general-purpose multimodal large language model: A survey,'' \emph{International Journal of Computer Vision}, pp. 1--39, 2025.

\bibitem{ouyang2022training}
L.~Ouyang, J.~Wu, X.~Jiang, D.~Almeida, C.~Wainwright, P.~Mishkin, C.~Zhang, S.~Agarwal, K.~Slama, A.~Ray \emph{et~al.}, ``Training language models to follow instructions with human feedback,'' \emph{Advances in neural information processing systems}, vol.~35, pp. 27\,730--27\,744, 2022.

\bibitem{vicuna2023}
\BIBentryALTinterwordspacing
W.-L. Chiang, Z.~Li, Z.~Lin, Y.~Sheng, Z.~Wu, H.~Zhang, L.~Zheng, S.~Zhuang, Y.~Zhuang, J.~E. Gonzalez, I.~Stoica, and E.~P. Xing, ``Vicuna: An open-source chatbot impressing gpt-4 with 90\%* chatgpt quality,'' March 2023. [Online]. Available: \url{https://lmsys.org/blog/2023-03-30-vicuna/}
\BIBentrySTDinterwordspacing

\bibitem{liu2024deepseek}
A.~Liu, B.~Feng, B.~Xue, B.~Wang, B.~Wu, C.~Lu, C.~Zhao, C.~Deng, C.~Zhang, C.~Ruan \emph{et~al.}, ``Deepseek-v3 technical report,'' \emph{arXiv preprint arXiv:2412.19437}, 2024.

\bibitem{radford2021learning}
A.~Radford, J.~W. Kim, C.~Hallacy, A.~Ramesh, G.~Goh, S.~Agarwal, G.~Sastry, A.~Askell, P.~Mishkin, J.~Clark \emph{et~al.}, ``Learning transferable visual models from natural language supervision,'' in \emph{International conference on machine learning}.\hskip 1em plus 0.5em minus 0.4em\relax PmLR, 2021, pp. 8748--8763.

\bibitem{zhai2023sigmoid}
X.~Zhai, B.~Mustafa, A.~Kolesnikov, and L.~Beyer, ``Sigmoid loss for language image pre-training,'' in \emph{Proceedings of the IEEE/CVF international conference on computer vision}, 2023, pp. 11\,975--11\,986.

\bibitem{li2023blip}
J.~Li, D.~Li, S.~Savarese, and S.~Hoi, ``Blip-2: Bootstrapping language-image pre-training with frozen image encoders and large language models,'' in \emph{International conference on machine learning}.\hskip 1em plus 0.5em minus 0.4em\relax PMLR, 2023, pp. 19\,730--19\,742.

\bibitem{brown2020language}
T.~Brown, B.~Mann, N.~Ryder, M.~Subbiah, J.~D. Kaplan, P.~Dhariwal, A.~Neelakantan, P.~Shyam, G.~Sastry, A.~Askell \emph{et~al.}, ``Language models are few-shot learners,'' \emph{Advances in neural information processing systems}, vol.~33, pp. 1877--1901, 2020.

\bibitem{team2023gemini}
G.~Team, R.~Anil, S.~Borgeaud, J.-B. Alayrac, J.~Yu, R.~Soricut, J.~Schalkwyk, A.~M. Dai, A.~Hauth, K.~Millican \emph{et~al.}, ``Gemini: a family of highly capable multimodal models,'' \emph{arXiv preprint arXiv:2312.11805}, 2023.

\bibitem{bai2023qwen}
J.~Bai, S.~Bai, Y.~Chu, Z.~Cui, K.~Dang, X.~Deng, Y.~Fan, W.~Ge, Y.~Han, F.~Huang \emph{et~al.}, ``Qwen technical report,'' \emph{arXiv preprint arXiv:2309.16609}, 2023.

\bibitem{touvron2023llama}
H.~Touvron, T.~Lavril, G.~Izacard, X.~Martinet, M.-A. Lachaux, T.~Lacroix, B.~Rozi{\`e}re, N.~Goyal, E.~Hambro, F.~Azhar \emph{et~al.}, ``Llama: Open and efficient foundation language models,'' \emph{arXiv preprint arXiv:2302.13971}, 2023.

\bibitem{li2024llavanext-strong}
\BIBentryALTinterwordspacing
B.~Li, K.~Zhang, H.~Zhang, D.~Guo, R.~Zhang, F.~Li, Y.~Zhang, Z.~Liu, and C.~Li, ``Llava-next: Stronger llms supercharge multimodal capabilities in the wild,'' May 2024. [Online]. Available: \url{https://llava-vl.github.io/blog/2024-05-10-llava-next-stronger-llms/}
\BIBentrySTDinterwordspacing

\bibitem{lu2024wildvision}
Y.~Lu, D.~Jiang, W.~Chen, W.~Y. Wang, Y.~Choi, and B.~Y. Lin, ``Wildvision: Evaluating vision-language models in the wild with human preferences,'' \emph{Advances in Neural Information Processing Systems}, vol.~37, pp. 48\,224--48\,255, 2024.

\bibitem{fu2024mmecomprehensiveevaluationbenchmark}
\BIBentryALTinterwordspacing
C.~Fu, P.~Chen, Y.~Shen, Y.~Qin, M.~Zhang, X.~Lin, J.~Yang, X.~Zheng, K.~Li, X.~Sun, Y.~Wu, and R.~Ji, ``Mme: A comprehensive evaluation benchmark for multimodal large language models,'' 2024. [Online]. Available: \url{https://arxiv.org/abs/2306.13394}
\BIBentrySTDinterwordspacing

\bibitem{liu2024mmbench}
Y.~Liu, H.~Duan, Y.~Zhang, B.~Li, S.~Zhang, W.~Zhao, Y.~Yuan, J.~Wang, C.~He, Z.~Liu \emph{et~al.}, ``Mmbench: Is your multi-modal model an all-around player?'' in \emph{European conference on computer vision}.\hskip 1em plus 0.5em minus 0.4em\relax Springer, 2024, pp. 216--233.

\bibitem{chen2021websrc}
X.~Chen, Z.~Zhao, L.~Chen, D.~Zhang, J.~Ji, A.~Luo, Y.~Xiong, and K.~Yu, ``Websrc: A dataset for web-based structural reading comprehension,'' \emph{arXiv preprint arXiv:2101.09465}, 2021.

\bibitem{liu2023hidden}
Y.~Liu, Z.~Li, H.~Li, W.~Yu, M.~Huang, D.~Peng, M.~Liu, M.~Chen, C.~Li, L.~Jin \emph{et~al.}, ``On the hidden mystery of ocr in large multimodal models,'' \emph{arXiv preprint arXiv:2305.07895}, vol.~2, no.~5, p.~6, 2023.

\bibitem{lu2023mathvista}
P.~Lu, H.~Bansal, T.~Xia, J.~Liu, C.~Li, H.~Hajishirzi, H.~Cheng, K.-W. Chang, M.~Galley, and J.~Gao, ``Mathvista: Evaluating mathematical reasoning of foundation models in visual contexts,'' \emph{arXiv preprint arXiv:2310.02255}, 2023.

\bibitem{zhang2024mathverse}
R.~Zhang, D.~Jiang, Y.~Zhang, H.~Lin, Z.~Guo, P.~Qiu, A.~Zhou, P.~Lu, K.-W. Chang, Y.~Qiao \emph{et~al.}, ``Mathverse: Does your multi-modal llm truly see the diagrams in visual math problems?'' in \emph{European Conference on Computer Vision}.\hskip 1em plus 0.5em minus 0.4em\relax Springer, 2024, pp. 169--186.

\bibitem{sun2023aligning}
Z.~Sun, S.~Shen, S.~Cao, H.~Liu, C.~Li, Y.~Shen, C.~Gan, L.-Y. Gui, Y.-X. Wang, Y.~Yang \emph{et~al.}, ``Aligning large multimodal models with factually augmented rlhf,'' \emph{arXiv preprint arXiv:2309.14525}, 2023.

\bibitem{li2023evaluating}
Y.~Li, Y.~Du, K.~Zhou, J.~Wang, W.~X. Zhao, and J.-R. Wen, ``Evaluating object hallucination in large vision-language models,'' \emph{arXiv preprint arXiv:2305.10355}, 2023.

\bibitem{maaz2023video}
M.~Maaz, H.~Rasheed, S.~Khan, and F.~S. Khan, ``Video-chatgpt: Towards detailed video understanding via large vision and language models,'' \emph{arXiv preprint arXiv:2306.05424}, 2023.

\bibitem{fu2025video}
C.~Fu, Y.~Dai, Y.~Luo, L.~Li, S.~Ren, R.~Zhang, Z.~Wang, C.~Zhou, Y.~Shen, M.~Zhang \emph{et~al.}, ``Video-mme: The first-ever comprehensive evaluation benchmark of multi-modal llms in video analysis,'' in \emph{Proceedings of the Computer Vision and Pattern Recognition Conference}, 2025, pp. 24\,108--24\,118.

\bibitem{pantic2005affective}
M.~Pantic, N.~Sebe, J.~F. Cohn, and T.~Huang, ``Affective multimodal human-computer interaction,'' in \emph{Proceedings of the 13th annual ACM international conference on Multimedia}, 2005, pp. 669--676.

\bibitem{hudlicka2003feel}
E.~Hudlicka, ``To feel or not to feel: The role of affect in human--computer interaction,'' \emph{International journal of human-computer studies}, vol.~59, no. 1-2, pp. 1--32, 2003.

\bibitem{gervasi2023applications}
R.~Gervasi, F.~Barravecchia, L.~Mastrogiacomo, and F.~Franceschini, ``Applications of affective computing in human-robot interaction: State-of-art and challenges for manufacturing,'' \emph{Proceedings of the Institution of Mechanical Engineers, Part B: Journal of Engineering Manufacture}, vol. 237, no. 6-7, pp. 815--832, 2023.

\bibitem{devillers2021human}
L.~Devillers, ``Human--robot interactions and affective computing: The ethical implications,'' in \emph{Robotics, AI, and humanity: Science, ethics, and policy}.\hskip 1em plus 0.5em minus 0.4em\relax Springer International Publishing Cham, 2021, pp. 205--211.

\bibitem{jin2024affective}
H.~Jin, C.~Qi, and Z.~Chen, ``Affective computing for healthcare: Recent trends, applications, challenges, and beyond,'' \emph{Emotional Intelligence}, p.~3, 2024.

\bibitem{yannakakis2018enhancing}
G.~N. Yannakakis, ``Enhancing health care via affective computing,'' 2018.

\bibitem{wu2016review}
C.-H. Wu, Y.-M. Huang, and J.-P. Hwang, ``Review of affective computing in education/learning: Trends and challenges,'' \emph{British Journal of Educational Technology}, vol.~47, no.~6, pp. 1304--1323, 2016.

\bibitem{yadegaridehkordi2019affective}
E.~Yadegaridehkordi, N.~F. B.~M. Noor, M.~N.~B. Ayub, H.~B. Affal, and N.~B. Hussin, ``Affective computing in education: A systematic review and future research,'' \emph{Computers \& education}, vol. 142, p. 103649, 2019.

\bibitem{lian2024affectgpt}
Z.~Lian, H.~Sun, L.~Sun, J.~Yi, B.~Liu, and J.~Tao, ``Affectgpt: Dataset and framework for explainable multimodal emotion recognition,'' \emph{arXiv preprint arXiv:2407.07653}, 2024.

\bibitem{zhang2025mme}
F.~Zhang, Z.~Cheng, C.~Deng, H.~Li, Z.~Lian, Q.~Chen, H.~Liu, W.~Wang, Y.-F. Zhang, R.~Zhang \emph{et~al.}, ``Mme-emotion: A holistic evaluation benchmark for emotional intelligence in multimodal large language models,'' \emph{arXiv preprint arXiv:2508.09210}, 2025.

\bibitem{lian2025emoprefer}
Z.~Lian, L.~Sun, L.~Chen, H.~Chen, Z.~Cheng, F.~Zhang, Z.~Jia, Z.~Ma, F.~Ma, X.~Peng \emph{et~al.}, ``Emoprefer: Can large language models understand human emotion preferences?'' \emph{arXiv preprint arXiv:2507.04278}, 2025.

\bibitem{yang2025humanomniv2}
Q.~Yang, S.~Yao, W.~Chen, S.~Fu, D.~Bai, J.~Zhao, B.~Sun, B.~Yin, X.~Wei, and J.~Zhou, ``Humanomniv2: From understanding to omni-modal reasoning with context,'' \emph{arXiv preprint arXiv:2506.21277}, 2025.

\bibitem{zhao2025r1}
J.~Zhao, X.~Wei, and L.~Bo, ``R1-omni: Explainable omni-multimodal emotion recognition with reinforcement learning,'' \emph{arXiv preprint arXiv:2503.05379}, 2025.

\bibitem{liu2023improved}
H.~Liu, C.~Li, Y.~Li, and Y.~J. Lee, ``Improved baselines with visual instruction tuning,'' 2023.

\bibitem{li2024llava}
B.~Li, Y.~Zhang, D.~Guo, R.~Zhang, F.~Li, H.~Zhang, K.~Zhang, P.~Zhang, Y.~Li, Z.~Liu \emph{et~al.}, ``Llava-onevision: Easy visual task transfer,'' \emph{arXiv preprint arXiv:2408.03326}, 2024.

\bibitem{Qwen2-VL}
P.~Wang, S.~Bai, S.~Tan, S.~Wang, Z.~Fan, J.~Bai, K.~Chen, X.~Liu, J.~Wang, W.~Ge, Y.~Fan, K.~Dang, M.~Du, X.~Ren, R.~Men, D.~Liu, C.~Zhou, J.~Zhou, and J.~Lin, ``Qwen2-vl: Enhancing vision-language model's perception of the world at any resolution,'' \emph{arXiv preprint arXiv:2409.12191}, 2024.

\bibitem{wang2025internvl3_5}
W.~Wang, Z.~Gao, L.~Gu, H.~Pu, L.~Cui, X.~Wei, Z.~Liu, L.~Jing, S.~Ye, J.~Shao \emph{et~al.}, ``Internvl3.5: Advancing open-source multimodal models in versatility, reasoning, and efficiency,'' \emph{arXiv preprint arXiv:2508.18265}, 2025.

\bibitem{qvq-72b-preview}
\BIBentryALTinterwordspacing
Q.~Team, ``Qvq: To see the world with wisdom,'' December 2024. [Online]. Available: \url{https://qwenlm.github.io/blog/qvq-72b-preview/}
\BIBentrySTDinterwordspacing

\bibitem{openaigpt4o}
\BIBentryALTinterwordspacing
OpenAI, ``Gpt-4o system card,'' 2024. [Online]. Available: \url{https://openai.com/index/gpt-4o-system-card}
\BIBentrySTDinterwordspacing

\bibitem{gemini25flash}
\BIBentryALTinterwordspacing
Google, ``Gemini 2.5 flash preview model card,'' 2025. [Online]. Available: \url{https://storage.googleapis.com/model-cards/documents/gemini-2.5-flash-preview.pdf}
\BIBentrySTDinterwordspacing

\end{thebibliography}

\end{document}